\documentclass[sigconf, nonacm]{acmart}

\usepackage{algorithmicx}
\usepackage[plain]{algorithm} 
\usepackage[noend]{algpseudocode}

\usepackage{xspace}

\usepackage{tikz}
\usepackage{pgfplots}
\usepackage{pgfplotstable}
\usepackage{subfigure}
\usetikzlibrary{patterns,shapes,arrows,positioning,fit,decorations.pathreplacing}
\usepackage{xcolor}
\usepgfplotslibrary{groupplots,patchplots}
\usepackage{dblfloatfix}
\usepackage[font=bf, skip=5pt]{caption}
\usetikzlibrary{shadows.blur}
\usetikzlibrary{shapes.geometric}
\usepackage{epigraph}

\newcommand{\algorithmicbreak}{\textbf{break}}

\algdef{SE}[FOR]{For}{EndFor}[1]{\algorithmicfor\ #1\ \textbf{:}}{\algorithmicend\ \algorithmicfor}%
\algdef{SE}[IF]{If}{EndIf}[1]{\algorithmicif\ #1\ \textbf{:}}{\algorithmicend\ \algorithmicif}%
\algdef{SE}[WHILE]{While}{EndWhile}[1]{\algorithmicwhile\ #1\ \textbf{:}}{\algorithmicend\ \algorithmicwhile}%

\makeatletter\ifthenelse{\equal{\ALG@noend}{t}}{\algtext*{EndFor}}{}\makeatother
\makeatletter\ifthenelse{\equal{\ALG@noend}{t}}{\algtext*{EndIf}}{}\makeatother
\makeatletter\ifthenelse{\equal{\ALG@noend}{t}}{\algtext*{EndWhile}}{}\makeatother

\theoremstyle{definition}
\newtheorem{definition}{Definition}

\newcommand{\rv}[1]{\mathbf{#1}}

\newcommand{\red}[1]{\textcolor{black}{#1}}

\newcommand*{\sys}{\textsc{Smart}\xspace}
\newcommand*{\sysa}{\textsc{Smart-ProfileAll}\xspace}
\newcommand*{\sysb}{\textsc{Smart-ProfileSmart}\xspace}
\newcommand*{\sysc}{\textsc{Smart-ModelMix}\xspace}

\newcommand\vldbdoi{XX.XX/XXX.XX}
\newcommand\vldbpages{XXX-XXX}
\newcommand\vldbvolume{14}
\newcommand\vldbissue{1}
\newcommand\vldbyear{2020}
\newcommand\vldbauthors{\authors}
\newcommand\vldbtitle{\shorttitle} 
\newcommand\vldbavailabilityurl{URL_TO_YOUR_ARTIFACTS}
\newcommand\vldbpagestyle{plain} 

\begin{document}

\title{SMART: Automatically Scaling Down Language Models with Accuracy Guarantees for Reduced Processing Fees}

\author{Saehan Jo}
\orcid{0000-0002-1825-0097}
\affiliation{%
  \institution{Cornell University}
  \city{Ithaca}
  \state{New York}
  \country{USA}
  \postcode{14850}
}
\email{sj683@cornell.edu}

\author{Immanuel Trummer}
\orcid{0000-0002-1825-0097}
\affiliation{%
  \institution{Cornell University}
  \city{Ithaca}
  \state{New York}
  \country{USA}
  \postcode{14850}
}
\email{itrummer@cornell.edu}

\begin{abstract}
The advancement of Large Language Models (LLMs) has significantly boosted performance in natural language processing (NLP) tasks. However, the deployment of high-performance LLMs incurs substantial costs, primarily due to the increased number of parameters aimed at enhancing model performance. This has made the use of state-of-the-art LLMs more expensive for end-users. AI service providers, such as OpenAI and Anthropic, often offer multiple versions of LLMs with varying prices and performance. However, end-users still face challenges in choosing the appropriate LLM for their tasks that balance result quality with cost.

We introduce \sys, \textbf{S}caling \textbf{M}odels \textbf{A}daptively for \textbf{R}educed \textbf{T}oken Fees, a novel LLM framework designed to minimize the inference costs of NLP tasks while ensuring sufficient result quality. It enables users to specify an accuracy constraint in terms of the equivalence of outputs to those of the most powerful LLM. \sys then generates results that deviate from the outputs of this LLM only with a probability below a user-defined threshold. \sys employs a profiling phase that evaluates the performance of multiple LLMs to identify those that meet the user-defined accuracy level. \sys optimizes the tradeoff between profiling overheads and the anticipated cost savings resulting from profiling. Moreover, our approach significantly reduces inference costs by strategically leveraging a mix of LLMs. Our experiments on three real-world datasets show that, based on OpenAI models, \sys achieves significant cost savings, up to 25.6$\times$ in comparison to GPT-4.
\end{abstract}

\maketitle

\pagestyle{\vldbpagestyle}
\begingroup\small\noindent\raggedright\textbf{PVLDB Reference Format:}\\
\vldbauthors. \vldbtitle. PVLDB, \vldbvolume(\vldbissue): \vldbpages, \vldbyear.\\
\href{https://doi.org/\vldbdoi}{doi:\vldbdoi}
\endgroup
\begingroup
\renewcommand\thefootnote{}\footnote{\noindent
This work is licensed under the Creative Commons BY-NC-ND 4.0 International License. Visit \url{https://creativecommons.org/licenses/by-nc-nd/4.0/} to view a copy of this license. For any use beyond those covered by this license, obtain permission by emailing \href{mailto:info@vldb.org}{info@vldb.org}. Copyright is held by the owner/author(s). Publication rights licensed to the VLDB Endowment. \\
\raggedright Proceedings of the VLDB Endowment, Vol. \vldbvolume, No. \vldbissue\ %
ISSN 2150-8097. \\
\href{https://doi.org/\vldbdoi}{doi:\vldbdoi} \\
}\addtocounter{footnote}{-1}\endgroup

\ifdefempty{\vldbavailabilityurl}{}{
\vspace{.3cm}
\begingroup\small\noindent\raggedright\textbf{PVLDB Artifact Availability:}\\
The source code, data, and/or other artifacts have been made available at \url{\vldbavailabilityurl}.
\endgroup
}

\section{Introduction}
\label{sec:intro}

\pgfplotscreateplotcyclelist{patternListIntro}{%
	{fill=black!50, postaction={pattern=horizontal lines}},
	{fill=brown!30, postaction={pattern=north west lines}},
	{fill=red!50, postaction={pattern=north east lines}},
	{fill=blue!50, postaction={pattern=crosshatch}},
	{fill=green!25, postaction={pattern=grid}},
	{fill=yellow!50, postaction={pattern=vertical lines}},
}

\begin{figure}
    \centering
    \begin{tikzpicture}
        \begin{groupplot}[group style={group size=1 by 1, x descriptions at=edge bottom, vertical sep=0cm}, width=6cm, height=3.8cm, ybar=0, ymode=log, xlabel={}, ylabel={Cost (\$)}, label style={font=\small, align=center}, typeset ticklabels with strut, xticklabels={IMDB}, xticklabel style={font=\small}, legend entries={gpt-4-0613, gpt-3.5-turbo-instruct, gpt-3.5-turbo-1106, davinci-002, babbage-002}, legend style={font=\small, legend columns=3}, xtick=data, ymajorgrids, ylabel near ticks, legend to name=groupPlotLegend, cycle list name=patternListIntro]
        \nextgroupplot[title={}, bar width=15pt, ymin=0.7, ytick={1, 10, 100, 1000}]
            \addplot table[skip first n=0, x expr=\coordindex, y index=0, col sep=comma] {plots/per_llm.csv};
            \addplot table[skip first n=0, x expr=\coordindex, y index=1, col sep=comma] {plots/per_llm.csv};
            \addplot table[skip first n=0, x expr=\coordindex, y index=2, col sep=comma] {plots/per_llm.csv};
            \addplot table[skip first n=0, x expr=\coordindex, y index=3, col sep=comma] {plots/per_llm.csv};
            \addplot table[skip first n=0, x expr=\coordindex, y index=4, col sep=comma] {plots/per_llm.csv};
        \end{groupplot}
    \end{tikzpicture}
    \ref{groupPlotLegend}
    \caption{Costs of OpenAI LLMs for the sentiment classification task on the IMDB benchmark.}
    \label{fig:intro}
\end{figure}
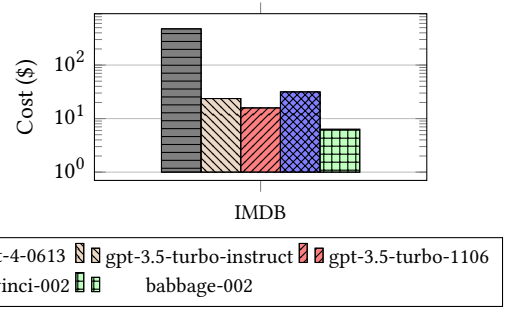

The rapid advancement and deployment of Large Language Models (LLMs) have started a new era of artificial intelligence (AI), significantly enhancing the capabilities of AI systems across a variety of natural language processing (NLP) tasks. Companies, such as OpenAI~\cite{OpenAI} and Anthropic~\cite{Anthropic}, have been at the forefront, offering LLMs as services to enable users to build intelligent, automated solutions. However, the deployment of these models, especially for a large number of input instances, comes with substantial costs. In recent years, there has been a notable trend in the development of LLMs towards increasing the number of parameters to boost their performance~\cite{abs-2001-08361, abs-2203-15556}. The expansion in model size has directly contributed to elevated inference costs, making it more expensive for end-users as well. 

Service providers often offer a variety of LLM options. For instance, OpenAI provides models such as GPT-4, GPT-3.5, davanci, and babbage~\cite{OpenAI}. Similarly, Google offers three sizes of its Gemini model: Ultra, Pro, and Nano~\cite{GoogleAI}. These LLMs exhibit heterogeneous performance and costs, in which a more powerful LLM is inevitably more expensive. The cost difference between these LLMs can span a couple of orders of magnitude, as illustrated in Figure~\ref{fig:intro}. However, for users, it is difficult to choose the right model for their NLP tasks, especially when there are no ground truth labels as a reference. Consequently, users tend to default to the most powerful LLM, resulting in unnecessary high costs. To improve the accuracy-cost tradeoff, the variation in costs and quality among LLMs warrants a strategic approach to selecting the most appropriate model for a given task. For that, we introduce \sys (\textbf{S}caling \textbf{M}odels \textbf{A}daptively for \textbf{R}educed \textbf{T}oken Fees), a framework aimed at improving the efficiency of inference while providing accuracy guarantees.

Our goal is to assist users in LLM inference, providing similar levels of output quality as the most powerful LLM but at a much lower expense. Primarily, we offer accuracy guarantees in terms of how closely the outputs of \sys match those of the highest-performing LLM, which we refer to as the reference LLM. \sys takes as input a task-specific question, a set of input instances, a user-defined accuracy constraint, and a confidence level. Given a collection of LLMs with varying performance and costs, \sys generates outputs that differ from those of the reference LLM only within the specified accuracy constraint and confidence level.

\sys comprises two main phases: profiling and application. The profiling phase aims to gather information on the accuracy of each LLM by comparing its outputs to those of the reference model. Once sufficient information is collected, \sys transitions to the application phase, where LLMs that meet (or closely approach) the accuracy criteria are used to process the remaining instances in a cost-efficient manner. Profiling carries a price because it involves processing the same item with multiple LLMs, including the reference model. Therefore, \sys employs a strategic approach to choose the right amount of profiling, balancing the costs of profiling against the anticipated savings in the application phase.

In the profiling phase, \sys gathers data on each LLM to assess its accuracy for the given NLP task. The aim is to identify LLMs whose outputs consistently align with those of the reference LLM, adhering to the user-defined accuracy constraint at the specified confidence level. To achieve this, \sys executes all available LLMs on the same input instance and compares their outputs with that of the reference LLM. \sys then iteratively accumulates more data by processing more items. This process is modeled as a series of Bernoulli trials, allowing \sys to calculate binomial proportion confidence intervals at the given confidence level for the accuracy of each LLM.  If the lower confidence interval bound of an LLM meets or exceeds the accuracy threshold, it is considered to have satisfied the accuracy constraint. This LLM can then be used in place of the reference LLM for subsequent LLM inferences during the application stage. Conversely, if the upper confidence interval bound falls below the accuracy threshold, it is deemed unsuitable for use. Profiling stops when the cheapest LLM that satisfies the accuracy constraint is found.

Profiling can be costly as it processes the same item with multiple LLMs. To address this, \sys employs an early termination criterion, halting the profiling phase early if additional profiling is anticipated to be wasteful. \sys assesses the benefit of additional profiling by weighing the extra costs of evaluating more items against the increased savings during the application phase. Estimating these expected costs is non-trivial, given that the true accuracy of each LLM remains unknown. \sys only observes samples of input instances, knowing how many of the processed items are identical to those of the reference LLM. Hence, \sys models the true accuracy as a binomial proportion under the binomial distribution. Based on this probabilistic approach, \sys calculates the expected cost of profiling additional items. If the anticipated costs of further profiling are projected to exceed its benefits, \sys halts the profiling process and moves on to the application phase.

A straightforward approach to the application phase involves employing the most affordable LLM that still meets the accuracy guarantees. However, further cost reductions can be achieved by leveraging a combination of multiple LLMs, even those whose individual accuracy does not meet the accuracy constraint. By strategically combining more affordable, less accurate LLMs with more expensive, more accurate ones, it is possible to further maximize cost savings. At the start of the application phase, \sys distributes the remaining inference workload over available LLMs to minimize the overall cost, while adhering to the user-defined accuracy constraint. \sys formulates this minimization problem as a mixed integer linear program (MILP), incorporating two key constraints: accuracy and confidence level. The accuracy constraint ensures that the collective accuracy across all inputs meets the given criteria. Similarly, the confidence level constraint requires that the confidence level aggregated from all deployed LLMs reaches the specified minimum. The solution to this MILP problem determines the ratios of the remaining input instances each LLM should process during the application phase.

In our experimental evaluation, we measure cost savings achieved by \sys on NLP tasks using three real-world datasets: IMDB~\cite{MaasDPHNP11}, SMS-Spam~\cite{AlmeidaHY11}, and AgNews~\cite{ZhangZL15}. We employ multiple OpenAI LLMs, including gpt-4-0613, gpt-3.5-turbo-instruct, gpt-3.5-turbo-1106, davinci-002, and babbage-002. Compared to GPT-4 (gpt-4-0613), across a variety of accuracy constraints, \sys demonstrates average cost savings of 7.2$\times$, 4.2$\times$, and 4.8$\times$ for the IMDB, SMS-Spam, and AgNews benchmarks, respectively. For instance, with an accuracy constraint of $\geq 90\%$, \sys achieves cost savings of 21.7$\times$, 16.0$\times$, and 21.8$\times$ for IMDB, SMS-Spam, and AgNews, respectively. In summary, the original contributions of this paper are as follows:

\begin{itemize}
\item We introduce \sys, a novel framework that minimizes the cost of LLM inference tasks while providing accuracy guarantees. This enables users to trade result quality for reduced costs.
\item We present a profiling scheme to evaluate the accuracy of LLMs in comparison to a reference LLM, along with an early termination criterion that effectively balances profiling overheads against future savings.
\item We detail our method for strategically leveraging multiple LLMs with varying performance and costs, maximizing cost savings while meeting an accuracy constraint.
\item We empirically evaluate \sys using real-world datasets and OpenAI LLMs, demonstrating significant cost savings and showcasing its ability to adapt to various accuracy targets.
\end{itemize}

The remainder of this paper is organized as follows. Section~\ref{sec:model} introduces our formal problem model and Section~\ref{sec:overview} gives an overview of the \sys framework. Section~\ref{sec:sysa} describes the profiling phase of \sys in more detail. Section~\ref{sec:sysb} describes the early termination criterion for profiling. Section~\ref{sec:sysc} describes the application phase of \sys. We report experimental results in Section~\ref{sec:experiment}. We discuss related work in Section~\ref{sec:related} before we conclude in Section~\ref{sec:conclusion}. The code of the \sys framework is available at \url{https://github.com/saehanjo/smart-llms}.

\section{Formal Model}
\label{sec:model}

\sys targets the following scenario.

\begin{definition}[LLMs with Varying Performance and Costs]
The performance of LLMs varies according to their sizes~\cite{abs-2001-08361, abs-2203-15556}. Service providers often offer a range of LLMs with varying sizes, leading to differences in performance and cost.
\end{definition}

\red{The costs of higher-performing LLMs increase substantially with their performance levels.} For instance, OpenAI charges $\$0.0015$ per input token for GPT-3.5 (turbo-0613), compared to $\$0.03$ for GPT-4 (0613), resulting in a cost difference of 20$\times$. Consequently, users find it challenging to decide which LLM to use for a specific task, especially in common scenarios where ground truth labels are not available as a benchmark. Using the most expensive LLM is the safest choice in terms of output quality but incurs significant costs. On the other hand, opting for \red{a cheaper and smaller} LLM might render the outputs unusable.


Using large language models is expensive. However, there are tasks where significantly smaller models generate almost equivalent output to larger ones. Our goal is to select the cheapest model that generates equivalent output to the reference model (i.e., the model whose output is assumed to be the most reliable, typically the most expensive model at the same time) with a probability above a user-defined threshold with a user-specified confidence level.


\begin{definition}[$\langle \delta, \gamma \rangle$-Equivalence]
\red{
Given a reference LLM $\overline{m}$, an LLM is $\langle \delta, \gamma \rangle$-equivalent if it generates outputs identical to those of the reference LLM with a probability of $1 - \delta$ at a confidence level of $\gamma$.
}
\end{definition}

\sys solves the following LLM inference problem.

\begin{definition}[Task-Specific LLM Inference with $\langle \delta, \gamma \rangle$-Equivalence Guarantees]
Given a question $q$, inputs $I$, a user-defined accuracy constraint $\delta$, and a confidence level $\gamma$, our goal is to generate outputs $O$ by leveraging LLMs $M$ with varying performance and costs. These outputs are equivalent to those of the reference LLM $\overline{m}$ in $100 \cdot (1 - \delta)\%$ of cases at a confidence level of $\gamma$.
\end{definition}
\section{\textsf{\textit{\sys}} Framework}
\label{sec:overview}

\begin{figure}[t]
\centering
\includegraphics[width=67mm]{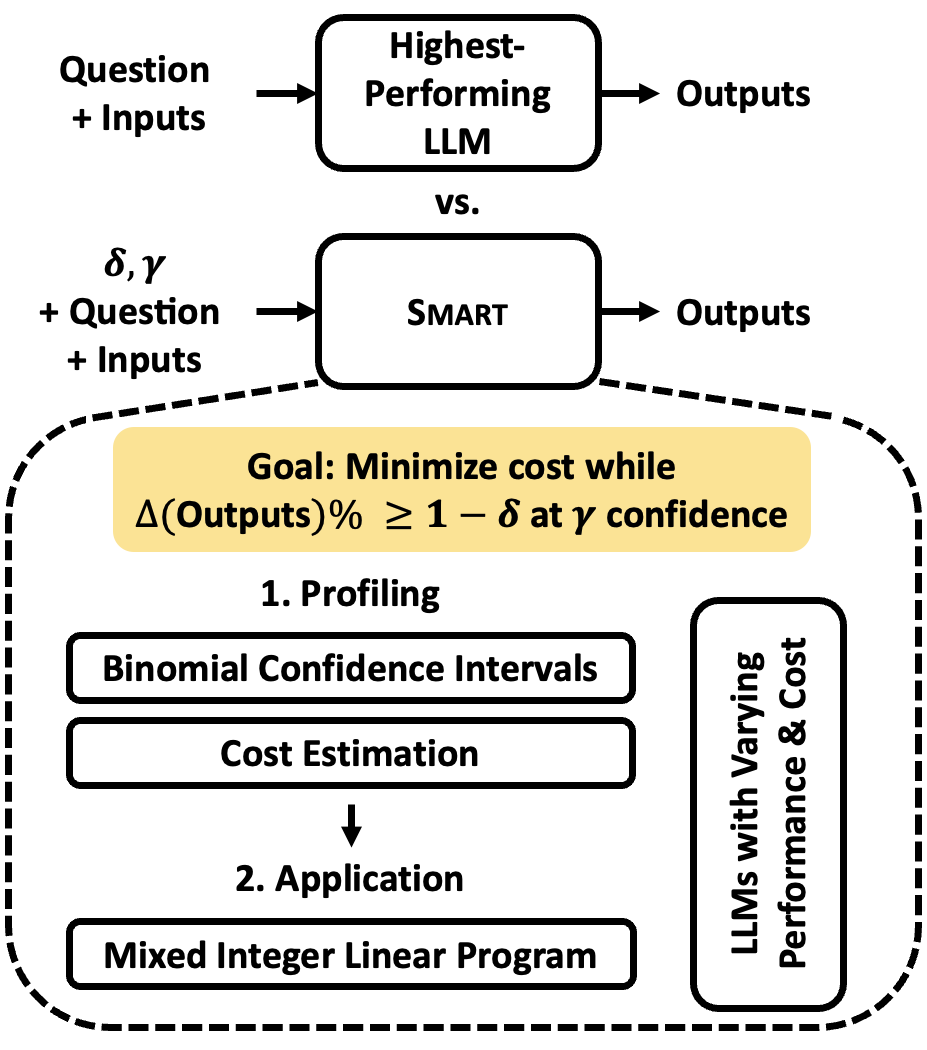}
\caption{\red{Overview of \sys.}}
\label{fig:overview}
\end{figure}

\begin{algorithm}[t]
    \small
    \begin{algorithmic}[1]
        \Statex \Comment{Generate outputs $O$ for a question $q$ from inputs $I$ using LLMs $M$, ensuring that at least $100 \cdot (1 - \delta)$\% of the outputs are consistent with those from the reference LLM $\overline{m}$ with a confidence level of $\gamma$.}
        \Function{\sys}{$M, \overline{m}, q, I, \delta, \gamma$}
        \Statex \hskip\algorithmicindent \Comment{Profile accuracy of LLMs.}
        \State $\langle M, O \rangle \gets \Call{Profile}{M, \overline{m}, q, I, \delta, \gamma}$
        \Statex \hskip\algorithmicindent \Comment{Calculate the ratio of items processed during profiling.}
        \State $r \gets |O|/|I|$
        \Statex \hskip\algorithmicindent \Comment{Remove items processed during profiling.}
        \State $I \gets \{i \in I | \nexists o:\langle i,o\rangle\in O\}$
        \Statex \hskip\algorithmicindent \Comment{Process remaining items based on LLM profiles.}
        \State $O \gets O\cup\Call{Apply}{M, q, I, \delta, \gamma, r}$
        \State \Return{$O$}
        \EndFunction
    \end{algorithmic}
    \caption{The \sys framework.\label{alg:canny}}
\end{algorithm}

Figure~\ref{fig:overview} presents an overview of the \sys framework. \sys is designed to minimize the cost of LLM inference on a large number of input items for a given task, while satisfying user-defined accuracy constraints. At its core, the framework features task-specific selection of LLMs with equivalence guarantees. We provide accuracy guarantees in terms of the equivalence of outputs compared to those of the reference model, which is typically the most powerful and expensive LLM. 

Algorithm~\ref{alg:canny} depicts a high-level overview of \sys. It takes in as input a question $q$, a set of inputs $I$, a reference LLM $\overline{m}$, a set of LLMs with varying costs $M$, a target accuracy $1 - \delta$, and a confidence level $\gamma$. \sys produces outputs $O$ for the question $q$ on inputs $I$, ensuring that the outputs differ from those of the reference model in no more than a $\delta$ ratio of cases with confidence $\gamma$. \sys consists of two phases: profiling and application. The main purpose of profiling is to estimate the model accuracy of each LLM with respect to the outputs of the reference LLM. Based on this information, we process the remaining items (after profiling) to maximize cost savings while satisfying the user-defined accuracy constraint. 

We present three variants of \sys, each adding a new feature to the preceding variant: 1) \sysa, 2) \sysb, and 3) \sysc. All versions of \sys share the same high-level structure, as shown in Algorithm~\ref{alg:canny}. The first version, \sysa, introduces the profiling phase, where we collect information on the accuracy of each available LLM. After profiling, it selects the most cost-efficient LLM, which is believed to meet the accuracy constraint with adequate confidence, to process the remaining items. The second version, \sysb, builds upon the first by evaluating the potential savings from profiling and by terminating profiling early when further profiling is considered wasteful. The final version, \sysc, enhances the application phase by selecting a combination of multiple models that together satisfy the user-defined accuracy constraints for processing the remaining items. We provide more details in the following sections.

\begin{table}
    \centering
    \small
    \caption{LLM $m$ and its attributes.\label{tbl:llm}}
    \begin{tabular}{lp{6.5cm}}
        \toprule[1pt]
        \textbf{Attribute} & \textbf{Semantics} \\
        \midrule[1pt]
        $m.n$ & Number of processed items using $m$ \\
        $m.e$ & Number of processed items whose outputs are equivalent to those of the reference model $\overline{m}$ \\
        $m.c$ & Average cost of processing a single item using $m$, which is updated as more items are processed \\
        $m.s$ & Status of $m$: \texttt{Unknown}, \texttt{Valid}, \texttt{Invalid} \\
        \bottomrule[1pt]
    \end{tabular}
\end{table}

\begin{algorithm}[t]
    \small
    \begin{algorithmic}[1]
        \Statex \Comment{Profile LLMs $M$ by evaluating whether their outputs for question $q$ from inputs $I$ are equivalent to those of the reference LLM $\overline{m}$ until sufficient information is obtained. The stopping criterion is based on the accuracy constraint $\delta$ and confidence level $\gamma$.}
        \Function{Profile}{$M, \overline{m}, q, I, \delta, \gamma$}
        \Statex \hskip\algorithmicindent \Comment{Initialize status of each LLM to unknown.}
        \State $\forall m \in M : m.s \gets \texttt{Unknown}$
        \Statex \hskip\algorithmicindent \Comment{Update status of the reference LLM to valid.}
        \State $\overline{m}.s \gets \texttt{Valid}$
        \Statex \hskip\algorithmicindent \Comment{Start profiling and store outputs of the reference LLM.}
        \State $O \gets \emptyset$
        \ForAll{$i \in I$}
            \Statex \hskip\algorithmicindent \hskip\algorithmicindent \Comment{Process item with the reference LLM $\overline{m}$.}
            \State $\overline{o} \gets \overline{m}(q, i)$
            \State $O \gets O \cup \{ \langle i, \overline{o} \rangle \}$
            \Statex \hskip\algorithmicindent \hskip\algorithmicindent \Comment{Process item with cheaper LLMs and check equivalence.}
            \ForAll{$m \in \{ m' \in M | m'.s = \texttt{Unknown} \}$}
                \State $m.n \gets m.n + 1$
                \State $o \gets m(q, i)$
                \If{$o = \overline{o}$}
                    \State $m.e \gets m.e + 1$
                \EndIf
            \EndFor
            \Statex \hskip\algorithmicindent \hskip\algorithmicindent \Comment{Update LLM status based on equivalence of their outputs.}
            \State $M \gets \Call{EvalModels}{M, \delta, \gamma}$
            \Statex \hskip\algorithmicindent \hskip\algorithmicindent \Comment{Number of remaining items to process.}
            \State $n \gets |\{i \in I | \nexists o:\langle i,o\rangle\in O\}|$
            \Statex \hskip\algorithmicindent \hskip\algorithmicindent \Comment{Stop profiling if condition is met.}
            \If{$\Call{TerminateProfile}{M, \overline{m}, \delta, \gamma, n}$}
                \State \algorithmicbreak
            \EndIf
            \EndFor
        \State \Return{$\langle M, O \rangle$}
        \EndFunction
    \end{algorithmic}
    \caption{Profile LLMs to find models satisfying the accuracy constraint.\label{alg:profile}}
\end{algorithm}

\begin{algorithm}[t]
    \small
    \begin{algorithmic}[1]
        \Statex \Comment{Given profiled LLMs $M$, terminate profiling if a valid LLM is as cheap as any LLM with unknown status.}
        \Function{TerminateProfile[ProfileAll]}{$M, \_, \_, \_, \_$}
        \Statex \hskip\algorithmicindent \Comment{Find valid LLM with the smallest cost.}
        \State $\underline{m} \gets \arg\min_{\{m \in M | m.s = \texttt{Valid}\}} m.c$
        \Statex \hskip\algorithmicindent \Comment{Terminate if a valid LLM is as cheap as any unknown LLM.}
        \If{$\underline{m}.c \leq \min_{\{m \in M | m.s = \texttt{Unknown}\}} m.c$}
            \State \Return{\texttt{True}}
        \EndIf
        \State \Return{\texttt{False}}
        \EndFunction
    \end{algorithmic}
    \caption{Terminate profiling if found cheapest LLM that satisfies the accuracy constraint.\label{alg:terminate}}
\end{algorithm}

\begin{algorithm}[t]
    \small
    \begin{algorithmic}[1]
        \Statex \Comment{Evaluate the status of each LLM $m\in M$ by calculating binomial confidence intervals with confidence $\gamma$, based on the number of processed items $m.n$ and the number of items with equivalent outputs $m.e$. An LLM is invalid if its accuracy upper bound falls below the accuracy threshold $1 - \delta$, and valid if its lower bound meets or exceeds the threshold.}
        \Function{EvalModels}{$M, \delta, \gamma$}
        \ForAll{$m \in \{ m' \in M | m'.s = \texttt{Unknown} \}$}
            \Statex \hskip\algorithmicindent \hskip\algorithmicindent \Comment{Compute binomial confidence intervals.}
            \State $\langle l, u \rangle \gets \Call{BinomCI}{m.n, m.e, \gamma}$
            \Statex \hskip\algorithmicindent \hskip\algorithmicindent \Comment{LLM is invalid if upper bound accuracy is too low.}
            \If{$u < 1 - \delta$}
                \State $m.s \gets \texttt{Invalid}$
            \EndIf
            \Statex \hskip\algorithmicindent \hskip\algorithmicindent \Comment{LLM is valid if lower bound accuracy is sufficiently high.}
            \If{$l \geq 1 - \delta$}
                \State $m.s \gets \texttt{Valid}$
            \EndIf
        \EndFor
        \State \Return{$M$}
        \EndFunction
    \end{algorithmic}
    \caption{Evaluate models as valid or invalid based on binomial confidence intervals.\label{alg:binomial}}
\end{algorithm}

\section{\textsf{\textit{\sysa}}: Profiling Models Exhaustively}
\label{sec:sysa}

The key feature of \sysa is the profiling phase. \sysa profiles all LLMs to determine whether each model meets or falls short of the user-defined accuracy threshold at the specified confidence level. For that, we introduce three possible status values for LLMs based on their current profile. A model is considered \texttt{Valid} if we have sufficient information to be confident that it meets the accuracy constraint. A model is deemed \texttt{Invalid} if we are confident that it fails to meet the accuracy constraint. The status of an LLM is \texttt{Unknown} if there is not enough information to either accept or reject the LLM as satisfying the accuracy constraint. \red{The profiling process ends when there is a \texttt{Valid} LLM as cost-efficient as any LLM with \texttt{Unknown} status. That is, we have identified the most affordable LLM that meets the accuracy constraint.}

Algorithm~\ref{alg:profile} illustrates the profiling phase in more detail. First, we initialize the status of each LLM (except the reference LLM) to \texttt{Unknown}. Next, we evaluate input items using the reference model as well as the other cheaper LLMs with \texttt{Unknown} status. We compare their outputs and keep track of the number of items processed and the number of items whose outputs are equivalent to those of the reference model. Then, we update the status of each LLM to either \texttt{Valid} or \texttt{Invalid} if we have gathered sufficient information using the \texttt{EvalModels} function, as presented in Algorithm~\ref{alg:binomial}. We provide more details in the following paragraph. Lastly, we stop profiling, when the termination criterion described in Algorithm~\ref{alg:terminate} is met. Profiling is terminated based on the current profiles of LLMs, specifically when there is a \texttt{Valid} LLM whose cost is lower than or equal to that of any LLM with \texttt{Unknown} status. This termination condition guarantees that we have identified the most cost-efficient LLM satisfying the user-defined accuracy constraint. Table~\ref{tbl:llm} summarizes the LLM profile properties, used in Algorithm~\ref{alg:profile} and the following algorithms.

Algorithm~\ref{alg:binomial} describes the process of evaluating the status of each LLM $m$ by calculating the binomial confidence interval for its accuracy. Here, we assume running an LLM on instances is a Bernoulli process, where the output of an instance could either match or differ from that of the reference model. The \textsc{BinomCI} function in Algorithm~\ref{alg:binomial} employs the Clopper-Pearson exact method~\cite{clopper1934use} to compute the binomial confidence intervals for LLM $m$. We denote the lower and upper bounds of the confidence interval by $l$ and $u$, respectively. If the upper bound of the model accuracy falls below the accuracy threshold (i.e., $u < 1 - \delta$), the model $m$ is considered \texttt{Invalid} for processing the remaining items. In contrast, if the lower bound of the model accuracy is higher than or equal to the accuracy threshold (i.e., $l \geq 1 - \delta$), then the status of $m$ is updated to \texttt{Valid}. If neither of these conditions is satisfied, we continue profiling $m$ in the subsequent iteration. \red{Figure~\ref{fig:confidence_interval} in the experimental section illustrates the development of confidence intervals on the accuracy of each LLM during the profiling phase.}

Algorithm~\ref{alg:single} outlines the application phase of \sysa. During this phase, we select the most cost-effective LLM among those identified as \texttt{Valid} (which may include the reference LLM). We then process each of the remaining items using the selected LLM. This procedure ensures that we leverage a cost-efficient solution while adhering to the accuracy guarantees established during the profiling phase. As described in the last line of Algorithm~\ref{alg:canny}, the outputs generated during this phase are combined with the outputs from the profiling phase and are presented to the user. \red{We denote different versions of algorithms for \sys variants using square brackets following the function name, as shown in Algorithm~\ref{alg:single}.}

We briefly discuss extensions to non-classification problems where a direct comparison of outputs (as in line 11 of Algorithm~\ref{alg:profile}) is challenging. For instance, for question answering, there is a low probability that the outputs of different LLMs will be exactly equal. One potential solution involves using a more sophisticated metric, such as a distance function comparing two texts, to assess the equivalence of the outputs. However, implementing a non-binary scoring function would require redesigning the framework, which currently relies on a Bernoulli process model. An effective alternative solution is to utilize the reference model to compare the two outputs for semantic matching. While this approach would increase the cost of profiling, it provides a binary metric that could seamlessly replace the exact output matching in line 11 of Algorithm~\ref{alg:profile}.

\begin{algorithm}[t]
    \small
    \begin{algorithmic}[1]
        \Statex \Comment{Given profiled LLMs $M$, question $q$, and remaining inputs $I$, process remaining items using the cheapest LLM that satisfies the accuracy constraint.}
        \Function{Apply[ProfileAll, ProfileSmart]}{$M, q, I, \_, \_, \_$}
        \Statex \hskip\algorithmicindent \Comment{Find cheapest LLM satisfying accuracy constraint.}
        \State $\underline{m} \gets \arg\min_{\{m \in M | m.s = \texttt{Valid}\}} m.c$
        \Statex \hskip\algorithmicindent \Comment{Process remaining items.}
        \State $O \gets \emptyset$
        \ForAll{$i \in I$}
            \State $o \gets \underline{m}(q, i)$
            \State $O \gets O \cup \{ \langle i, o \rangle \}$
        \EndFor
        \State \Return{$O$}
        \EndFunction
    \end{algorithmic}
    \caption{Single-model application.\label{alg:single}}
\end{algorithm}

\begin{algorithm}[t]
    \small
    \begin{algorithmic}[1]
        \Statex \Comment{Given profiled LLMs $M$, reference LLM $\overline{m}$, accuracy constraint $\delta$, confidence level $\gamma$, and the remaining $n$ items, terminate profiling if further profiling is expected to be wasteful (or a valid LLM is as cheap as any LLM with unknown status).}
        \Function{TerminateProfile[ProfileSmart]}{$M, \overline{m}, \delta, \gamma, n$}
        \Statex \hskip\algorithmicindent \Comment{Terminate if a valid LLM is as cheap as any unknown LLM.}
        \If{$\Call{TerminateProfile[ProfileAll]}{M}$}
            \State \Return{\texttt{True}}
        \EndIf
        \Statex \hskip\algorithmicindent \Comment{Find current cheapest LLM satisfying accuracy constraint.}
        \State $\underline{m} \gets \arg\min_{\{m \in M | m.s = \texttt{Valid}\}} m.c$
        \Statex \hskip\algorithmicindent \Comment{Compute the expected cost if profiling terminates at this point.}
        \State $\underline{c} \gets n \cdot \underline{m}.c$
        \Statex \hskip\algorithmicindent \Comment{Store expected costs for varying numbers of profiled items.}
        \State $k = 1$
        \While{$k \leq n$}
            \Statex \hskip\algorithmicindent \hskip\algorithmicindent \Comment{Compute the expected cost if profiling exactly $k$ more items.}
            \State $c_k \gets \Call{Cost}{k, M, \overline{m}, \underline{m}, \delta, \gamma, n}$
            \State $k \gets 2 \cdot k$
        \EndWhile
        \Statex \hskip\algorithmicindent \Comment{Terminate if further profiling is expected to be wasteful.}
        \If{$\underline{c} \leq \min_k c_k$}
            \State \Return{\texttt{True}}
        \EndIf
        \State \Return{\texttt{False}}
        \EndFunction
    \end{algorithmic}
    \caption{Terminate profiling if further profiling is expected to be wasteful.\label{alg:terminate_expected}}
\end{algorithm}

\section{\textsf{\textit{\sysb}}: Restricting Profiling Overheads}
\label{sec:sysb}

\sysb introduces a feature that terminates profiling early if further evaluation is expected to be wasteful. Profiling involves running both the reference LLM, which is typically the most expensive, and the other LLMs on the same input instance. Our hope is that the cost of using the reference LLM is offset by the savings from employing a less expensive LLM for processing the remaining items after profiling. However, profiling costs can outweigh the cost savings when a large number of items are needed to validate an LLM with \texttt{Unknown} status as either \texttt{Valid} or \texttt{Invalid}. Therefore, we estimate the expected cost savings from profiling additional items and terminate early if the projected savings are not positive.

Algorithm~\ref{alg:terminate_expected} illustrates the newly added termination criterion based on expected cost savings. First, we calculate the expected cost of using the most cost-efficient LLM among the current set of \texttt{Valid} LLMs to process the remaining items. For that, we multiply the number of remaining items by the unit cost (refer to $m.c$ in Table~\ref{tbl:llm}) of the cheapest \texttt{Valid} LLM. This serves as a baseline, assuming we cease profiling at this juncture and proceed to the application phase. Next, we assess the impact of profiling more items on the overall cost. Specifically, we calculate the expected costs for different values of $k$, where $k$ represents the number of additionally profiled items. We do not know a priori which number of profiled items will yield the optimal balance between profiling overheads and the cost savings in the application phase. Hence, we start with $k=1$ and incrementally double its value up to the number of remaining items. This exponential scheme is reasonable, given that profiling too many items would be wasteful, and the search should concentrate on the range with smaller values. Finally, we compare the lowest among the costs of profiling additional items against the cost of halting profiling at this point. We terminate profiling if setting $k = 0$ minimizes expected costs.

The \textsc{Cost} function in Algorithm~\ref{alg:terminate_expected} calculates the expected cost of profiling LLMs for exactly $k$ more items, followed by the application phase. That is, we continue profiling for $k$ additional items and then process remaining items based on the newly collected information. To compute the expected cost, we first need to determine for each LLM $m$ with \texttt{Unknown} status (i.e., $m.s = \texttt{Unknown}$) whether $m$ will eventually be considered \texttt{Valid} after profiling for $k$ more items. We express the probability that LLM $m$ will be evaluated as \texttt{Valid} as the following:
\begin{equation*}
\Pr(m.s=\texttt{Valid} | k, \delta, \gamma)
\end{equation*}

Next, we assign a sort order to LLMs based on their unit costs (since the same prompt is used for all LLMs, this effectively amounts to ordering the LLMs based on their per-token costs). Among the LLMs with \texttt{Unknown} status, we label the LLM with the smallest unit cost as $m_1$, the one with the second smallest unit cost as $m_2$, and so on. Based on this ordering, if $m_1$ is evaluated to be \texttt{Valid}, it will be used during the application phase due to its smallest unit cost. If $m_1$ is not \texttt{Valid}, then the LLM $m_2$ with the next smallest unit cost will be considered. Namely, this ordering represents the sequence in which the LLMs will be considered during the application stage. Lastly, \red{assuming independence between LLMs}, we can formulate the expected cost as follows. For simplicity, we denote the probability that LLM $m_i$ is evaluated as \texttt{Valid} by $p_i = \Pr(m_i.s=\texttt{Valid} | k, \delta, \gamma)$.
\begin{equation}
\label{eq:expected_cost}
\begin{split}
\Call{Cost}{k, M, \overline{m}, \underline{m}, \delta, \gamma, n} & = k ( \overline{m}.c + \sum_i m_i.c ) \\
& + (n - k) \sum_i \bigl( \prod_{j < i} (1 - p_j) \bigr) \cdot p_i \cdot m_i.c \\
& + (n - k) \bigl( \prod_i (1 - p_i) \bigr) \cdot \underline{m}.c \\
\end{split}
\end{equation}

The first term, $k ( \overline{m}.c + \sum_i m_i.c )$, represents the cost of profiling $k$ additional items using the reference LLM $\overline{m}$ alongside the LLMs with \texttt{Unknown} status $\{ m_i \}$. The second term accounts for the cost of processing the remaining $(n - k)$ items using the cheapest LLM among those newly verified as \texttt{Valid}. The third term addresses the scenario where none of the LLMs are evaluated to be \texttt{Valid} and thus rendering the profiling wasteful. By summing these terms together, we derive the expected cost of profiling exactly $k$ more items and processing the remaining items in the application phase.

To calculate the expected cost formula, we need to determine the probability of an LLM being \texttt{Valid} after profiling $k$ more items. An LLM is evaluated as \texttt{Valid} if the lower bound of the binomial confidence interval on its accuracy is greater than or equal to the accuracy constraint $1 - \delta$ with confidence $\gamma$. Thus, we obtain:
\begin{equation*}
\Pr(m.s=\texttt{Valid} | k, \delta, \gamma) = \Pr(l_m \geq 1 - \delta)
\end{equation*}
where $l_m$ is the lower bound of the binomial confidence interval (with confidence $\gamma$) for LLM $m$ after profiling $k$ additional items. We now want to compute the binomial confidence interval after profiling $k$ more items. Recall that profiling an LLM $m$ reveals that after $m.n$ items are processed, $m.e$ items produce outputs equivalent to the reference model. Based on this current information, the number of processed items is calculated as the sum of $k$ and the number of currently profiled items $m.n$. For LLM $m$, among the $k$ newly profiled items, we denote the number of items with the same outputs as the reference model by $e_m$. Similar to the number of processed items, the number of conforming items is given as the sum of $e_m$ and $m.e$. The confidence level is given by $\gamma$. Thus, the lower bound $l_m$ of the binomial confidence interval is derived from the function: \Call{BinomCI}{$(k + m.n), (e_m + m.e), \gamma$}. As a result, we can solve the inequality $l_m \geq 1 - \delta$ for $e_m$ to find the smallest value of $e_m$ (which we denote by $e_m^*$) that satisfies the following condition:
\begin{equation*}
\Call{BinomCI}{(k + m.n), (e_m + m.e), \gamma}.\textsc{Lower} \geq 1 - \delta
\end{equation*}

By treating the number of conforming items $e_m$ as a random variable $\rv{e}$, we rewrite the probability of an LLM being \texttt{Valid} after profiling $k$ additional items as:
\begin{equation*}
\Pr(m.s=\texttt{Valid} | k, \delta, \gamma) = \Pr(l_m \geq 1 - \delta) = \Pr(\rv{e} \geq e_m^*)
\end{equation*}
Random variables are represented by symbols in bold font. If given the true accuracy of LLM $m$ when evaluated on all items as $a$, we can compute this probability using the binomial distribution. The random variable $\rv{e}$ follows the binomial distribution with parameters $k$ and $a$ as: $\rv{e} \sim \Call{Binom}{k, a}$. Therefore, by denoting the cumulative distribution function (CDF) of this binomial distribution as $B(x) = \Pr(\rv{e} \leq x)$, we get the probability as:
\begin{equation*}
\Pr(\rv{e} \geq e_m^*|a) = 1 - B(e_m^* - 1)
\end{equation*}

However, the challenge is that we do not know the true accuracy $a$ of LLM $m$. We instead rely on the observations from profiling, $m.n$ and $m.e$, to estimate the true accuracy as $\hat{a} = m.e / m.n$. Then, under the same assumptions as those used in the Wald interval (based on the central limit theorem)~\cite{brown2001interval}, we approximate the estimation error using a Gaussian distribution as follows:
\begin{equation*}
\frac{\hat{a} - a}{\sqrt{\hat{a} (1 - \hat{a}) / m.n}} \sim \Call{Norm}{0, 1}
\end{equation*}
That is, the random variable $\rv{a}$, which models the true accuracy $a$, follows a Gaussian distribution as the following:
\begin{equation*}
\rv{a} \sim \Call{Norm}{\mu=\hat{a}, \sigma^2=\frac{\hat{a} (1 - \hat{a})}{m.n}}
\end{equation*}
The intuition behind this formalization is that as we collect more information (i.e., profile more items), the variance of the error decreases due to the term $\frac{1}{m.n}$.  In other words, as the sample size increases, it becomes more likely that the sample mean $\hat{a}$ accurately represents the true accuracy $a$.

Finally, we compute the probability of LLM $m$ being \texttt{Valid} after processing $k$ additional items by integrating over the possible range of $a \in [0, 1]$. By denoting the probability density function (PDF) of this normal distribution as $f(x) = \Pr(\rv{a} = x)$, we get the following formula:
\begin{equation*}
\begin{split}
\Pr(\rv{e} \geq e_m^*) & = \int_0^1 \Pr(\rv{e} \geq e_m^* | a) \Pr(a) \,da \\
& = \int_0^1  (1 - B(e_m^* - 1)) f(a) \,da \\
\end{split}
\end{equation*}
where $B(x)$ is the CDF of $\Call{Binom}{k, a}$ and $f(x)$ is the PDF of $\Call{Norm}{\hat{a}, \frac{\hat{a} (1 - \hat{a})}{m.n}}$. We can now compute $p_i$ in Equation~\ref{eq:expected_cost} and use the function to calculate expected costs. Figures~\ref{fig:expected_cost} and \ref{fig:expected_cost_inverse} in the experimental section illustrate the expected costs of profiling additional items, including the profiling overheads and the savings anticipated in the application phase.

The application phase of \sysb is identical to that of \sysa. As described in Algorithm~\ref{alg:single}, we select the \texttt{Valid} LLM with the lowest unit cost and apply it to the remaining items. The distinction between the two variants of \sys lies in the profiling phase, where \sysb carefully evaluates the tradeoff between profiling overheads and application savings.

\begin{algorithm}[t]
    \small
    \begin{algorithmic}[1]
        \Statex \Comment{Given profiled LLMs $M$, question $q$, remaining inputs $I$, and the ratio of items processed during profiling $r$, process remaining items using multiple LLMs to minimize cost while satisfying the accuracy constraint $\delta$ with a confidence level $\gamma$.}
        \Function{Apply[ModelMix]}{$M, q, I, \delta, \gamma, r$}
        \Statex \hskip\algorithmicindent \Comment{Find ratio per LLM.}
        \State $\{ r_m \}_{m\in M} \gets \Call{ComputeRatios}{M, \delta, \gamma, r}$
        \Statex \hskip\algorithmicindent \Comment{Split remaining items based on ratios.}
        \State $\{ I_m \}_{m\in M} \gets \Call{PartitionByRatios}{I, \{ r_m \}}$
        \Statex \hskip\algorithmicindent \Comment{Process remaining items.}
        \State $O \gets \emptyset$
        \ForAll{$m \in M$}
            \ForAll{$i \in I_m$}
                \State $o \gets m(q, i)$
                \State $O \gets O \cup \{ \langle i, o \rangle \}$
            \EndFor
        \EndFor
        \State \Return{$O$}
        \EndFunction
    \end{algorithmic}
    \caption{Multiple-models application.\label{alg:multi}}
\end{algorithm}

\section{\textsf{\textit{\sysc}}: Selecting Model Combinations}
\label{sec:sysc}

\sysc improves the application phase by leveraging all LLMs to maximize cost savings. Previously, in Algorithm~\ref{alg:single}, we employ a single LLM to process the remaining items in the application phase. This earlier method misses a potential opportunity for additional cost savings. Consider an LLM that is more cost-efficient but has an accuracy lower bound just below the accuracy threshold. Such slight shortfall in accuracy disqualifies the LLM as a valid option for application in the previous method. However, recognizing that its accuracy is close to the user-defined threshold, \sysc proposes a novel approach. The main idea involves combining this LLM with an LLM with a higher accuracy. By partitioning the processing of remaining items between the more economical LLM and a more expensive, higher-accuracy LLM, we can realize further cost reductions.

Algorithm~\ref{alg:multi} describes the novel approach for the application phase. At its heart is the \texttt{ComputeRatio} function, which determines the percentage of remaining items that each LLM should process. This calculation is based on solving a mixed integer linear program, with further details provided in the subsequent paragraphs. We then partition the remaining items according to the determined ratios and process them using the respective LLMs. Due to this careful partitioning, we ensure that the generated outputs, in aggregate, meet the accuracy constraint at the designated confidence level.

We provide a detailed explanation of the \texttt{ComputeRatio} function. Our objective is to minimize the cost of processing the remaining items while adhering to the accuracy constraint. To achieve this, we focus on minimizing the cost function below, where $i$ indexes all available LLMs:
\begin{equation*}
\min_i c_i x_i
\end{equation*}
where $c_i$ is the unit cost of an LLM $m_i$ and $x_i \in [0, 1]$ is the ratio of items processed by model $m_i$. The ratios $x_i$ should sum up to one: $\sum_i x_i = 1$.

There are two key constraints associated with this minimization problem: 1) accuracy constraint and 2) confidence level constraint. We need to guarantee that, after processing all items, we satisfy the accuracy threshold $1 - \delta$ with confidence $\gamma$. To specify these constraints, we first introduce a separate confidence level $\gamma_i$ for each LLM $m_i$. The main idea is that the combined confidence of these individual confidence levels $\gamma_i$ should be greater than or equal to the confidence constraint $\gamma$. As a result, assuming independence between confidence levels, we get the confidence level constraint as follows:
\begin{equation}
\label{eq:confidence}
\prod_i \gamma_i \geq \gamma
\end{equation}
Here, without loss of generality, we say that LLMs (except for the reference model) exhibit zero accuracy lower bound for a 100\% confidence level.

Then, we specify the accuracy constraint based on the accuracy lower bound of each LLM $m_i$. First, we compute the lower bound $l_i$ on the accuracy of LLM $m_i$, using the binomial confidence intervals with confidence $\gamma_i$. That is, for each LLM $m_i$, given the number of processed items, $m_i.n$, and the number of items whose outputs conform with the reference LLM, $m_i.e$, we calculate the lower bound from the function \Call{BinomCI}{$m_i.n, m_i.e, \gamma_i$}. Now, we need to guarantee that the final accuracy satisfies the given accuracy constraint. In other words, we want the accumulated accuracy over all items to be at least the same as the accuracy threshold. \red{We temporarily denote this accuracy threshold by $\alpha$ instead of the previously used $1 - \delta$ (since its value slightly differs from $1 - \delta$, a distinction we will explain shortly).} Given the accuracy lower bound $l_i$ per LLM $m_i$, we specify the accuracy constraint as:

\begin{equation}
\label{eq:accuracy}
\sum_i l_i x_i \geq \alpha
\end{equation}

When computing the final accuracy over all items, we should also take into account the items that we processed during the profiling phase (which we processed using the reference LLM). In that, we use a slightly refined version of the accuracy threshold (instead of $\alpha = 1 - \delta$) in this optimization problem. Note that the accuracy of the reference LLM is one, as we define accuracy in terms of consistency with the outputs of the reference model. Denoting by $r$ the ratio of items processed in the profiling phase to the total number of items, we solve the following equation to get the refined accuracy threshold $\alpha$:
\begin{equation*}
1 \cdot r + \alpha (1 - r)  = 1 - \delta \Rightarrow \alpha = 1 - \frac{\delta}{1 - r}
\end{equation*}

We solve this minimization problem for $x_i$ to find the optimal ratios for distributing the remaining items to LLMs. However, solving this problem directly is difficult due to the non-elementary natural of computing quantiles (e.g., Gauss error function if using a normal approximation). That is, it is hard to express the accuracy lower bounds $l_i$ as an elementary function of other variables. Instead, we define a fixed set of potential confidence levels $\{ \gamma_j \}$ where each level is indexed by $j$ (instead of $i$). Then, we introduce a binary variable $y_{ij} \in \{ 0, 1 \}$ that indicates whether the LLM $m_i$ operates under the confidence level $\gamma_j$. We assign at most one confidence level per LLM, thereby introducing the following constraint: $\forall i: \sum_j y_{ij} \leq 1$. In our implementation, we utilize a range of confidence levels starting from the user-defined confidence level $\gamma$ up to one, with an increment of $0.01$. For instance, if a user specifies a confidence level of $0.95$, we consider potential confidence levels $\gamma_j \in \{ 0.95, 0.96, 0.97, 0.98, 0.99, 1 \}$. Now, we reformulate the two constraints in Equations~\ref{eq:confidence} and \ref{eq:accuracy}.

First, we rewrite the accuracy constraint in Equation~\ref{eq:accuracy}. Since we specify predefined values $\{ \gamma_j \}$ for confidence levels, we can compute the accuracy lower bound $l_{ij}$ for each combination of LLM $m_i$ and confidence level $\gamma_j$. Now, we replace the term $l_i$ in Equation~\ref{eq:accuracy} using the new variables: $l_i = \sum_j l_{ij} y_{ij}$. Note that, due to the constraint on $y_{ij}$, only one of $y_{ij}$ can have a value of one for a fixed $i$. As a result, we reformulate Equation~\ref{eq:accuracy} as the following:
\begin{equation}
\label{eq:acc_modified}
\sum_{ij} l_{ij} x_i y_{ij} \geq 1 - \frac{\delta}{1 - r}
\end{equation}

Second, we rewrite the confidence level constraint by replacing $\gamma_i$ using the new variables $\gamma_j$ and $y_{ij}$. The confidence level $\gamma_i$ of LLM $m_i$ can be expressed as: $\gamma_i = \prod_j (1 - (1 - \gamma_j) y_{ij})$. The term inside the product has a value of one when $y_{ij} = 0$ and otherwise $\gamma_j$ when $y_{ij} = 1$. Thus, we reformulate Equation~\ref{eq:confidence} as follows:
\begin{equation}
\label{eq:conf_product}
\prod_{ij} (1 - (1 - \gamma_j) y_{ij}) \geq \gamma
\end{equation}

In order to formulate this minimization problem as a mixed integer linear program (MILP), we convert the product in Equation~\ref{eq:conf_product} into a summation. By applying logarithmic function on both sides, we obtain:
\begin{equation*}
\sum_{ij} \ln (1 - (1 - \gamma_j) y_{ij}) \geq \ln \gamma
\end{equation*}

To extract the binary variables $y_{ij}$ from the logarithmic function, we use the Taylor expansion~\cite{abramowitz1948handbook} of the form: $\ln (1 - x) = - \sum_{n=1}^{\infty} \frac{x^n}{n} = A_\infty (x)$. Here, $A_n (x)$ is the polynomial approximation of degree $n$. We set $x = (1 - \gamma_j) y_{ij}$ and simplify the function $A_\infty ((1 - \gamma_j) y_{ij})$ as the following:
\begin{equation*}
A_\infty ((1 - \gamma_j) y_{ij}) = \sum_{n=1}^{\infty} \frac{((1 - \gamma_j) y_{ij})^n}{n} = \sum_{n=1}^{\infty} \frac{(1 - \gamma_j)^n}{n}  (y_{ij})^n 
\end{equation*}
Since $y_{ij} \in \{0, 1\}$ is a binary variable, we get $(y_{ij})^n = y_{ij}$. We further simplify the function as follows:
\begin{equation*} 
\begin{split}
\sum_{n=1}^{\infty} \frac{(1 - \gamma_j)^n}{n}  (y_{ij})^n & = \sum_{n=1}^{\infty} \frac{(1 - \gamma_j)^n}{n}  y_{ij} = y_{ij} A_\infty (1 - \gamma_j) \\
& = y_{ij} \ln (1 - (1 - \gamma_j)) = y_{ij} \ln \gamma_j
\end{split}
\end{equation*}
Therefore, we rewrite the confidence level constraint as the following:
\begin{equation}
\label{eq:conf_modified}
\sum_{ij} y_{ij} \ln \gamma_j \geq \ln \gamma
\end{equation}

In summary, based on the reformulated constraints in Equations~\ref{eq:acc_modified} and \ref{eq:conf_modified}, we solve the following MILP problem:
\begin{equation*}
\min_i c_i x_i
\end{equation*}
subject to
\begin{equation*}
\sum_{ij} l_{ij} x_i y_{ij} \geq 1 - \frac{\delta}{1 - r} ,
\end{equation*}
\begin{equation*}
\sum_{ij} y_{ij} \ln \gamma_j \geq \ln \gamma ,
\end{equation*}
\begin{equation*}
\sum_i x_i = 1 ,
\end{equation*}
and
\begin{equation*}
\forall i: \sum_j y_{ij} \leq 1
\end{equation*}
where $x_i \in [0, 1]$ and $y_{ij} \in \{ 0, 1 \}$.

\section{Experimental Evaluation}
\label{sec:experiment}

\pgfplotscreateplotcyclelist{patternList}{%
    {black,fill=gray,postaction={pattern=horizontal lines}},%
    {brown!60!black,fill=brown!30!white,postaction={pattern=north west lines}},%
    {red,fill=red!30!white,postaction={pattern=north east lines}},%
    {blue,fill=blue!30!white,postaction={pattern=crosshatch}},%
    {green,fill=green!80!black,postaction={pattern=vertical lines}}%
    {violet!80!black,fill=violet,postaction={pattern=grid}},%
}
\pgfplotscreateplotcyclelist{patternList2}{%
	{fill=black!50, postaction={pattern=horizontal lines}},
	{fill=brown!30, postaction={pattern=north west lines}},
	{fill=red!50, postaction={pattern=north east lines}},
	{fill=blue!50, postaction={pattern=crosshatch}},
	{fill=green!25, postaction={pattern=grid}},
	{fill=yellow!50, postaction={pattern=vertical lines}},
}

We provide the experimental setup in Section~\ref{sec:exp_setup} and present the experimental results in Section~\ref{sec:exp_result}.


\begin{table}
    \centering
    \small
    \caption{Overview of benchmarks.\label{tbl:benchmarks}}
    \begin{tabular}{lrrr}
        \toprule[1pt]
        \textbf{Dataset} & \textbf{\#Instances} & \textbf{\#Labels} & \textbf{\#Tokens per Instance} \\
        \midrule[1pt]
        IMDB~\cite{MaasDPHNP11} & 50,000 & 2 & 293.7 \\
        SMS-Spam~\cite{AlmeidaHY11} & 5,574 & 2 & 22.9 \\
        AgNews~\cite{ZhangZL15} & 127,600 & 4 & 51.2 \\
        \bottomrule[1pt]
    \end{tabular}
\end{table}

\begin{figure}
    \centering
    \begin{tikzpicture}
        \begin{groupplot}[group style={group size=1 by 1, x descriptions at=edge bottom, vertical sep=1.40cm}, width=8.8cm, height=3.8cm, ybar=0, xlabel={}, ylabel={}, label style={font=\small, align=center}, nodes near coords={\ifnum\coordindex=1{\pgfmathprintnumber{\pgfplotspointmeta}}\fi}, nodes near coords style={/pgf/number format/.cd, fixed zerofill, precision=1}, enlarge x limits=0.3, typeset ticklabels with strut, xticklabels={IMDB, SMS-Spam, AgNews}, xticklabel style={font=\small}, legend entries={GPT-4, \sysa, \sysb, \sys (i.e.\, \sysc)}, legend style={font=\small, at={(0.45,-0.41)}, anchor=north, legend columns=2}, xtick=data, ymajorgrids, ylabel near ticks, legend to name=groupPlotLegend, cycle list name=patternList]
        \nextgroupplot[title={}, bar width=15pt, ylabel={Cost (\$)}, ytick={0, 100, 200, 300, 400, 500}]
            \addplot table[skip first n=0, x expr=\coordindex, y index=1, col sep=comma] {plots/aggregated.csv};
            \addplot table[skip first n=0, x expr=\coordindex, y index=2, col sep=comma] {plots/aggregated.csv};
            \addplot table[skip first n=0, x expr=\coordindex, y index=3, col sep=comma] {plots/aggregated.csv};
            \addplot table[skip first n=0, x expr=\coordindex, y index=4, col sep=comma] {plots/aggregated.csv};
        \end{groupplot}
    \end{tikzpicture}
    \ref{groupPlotLegend}
    \caption{Costs of GPT-4 and all variants of \sys on all benchmarks.}
    \label{fig:aggregated}
\end{figure}
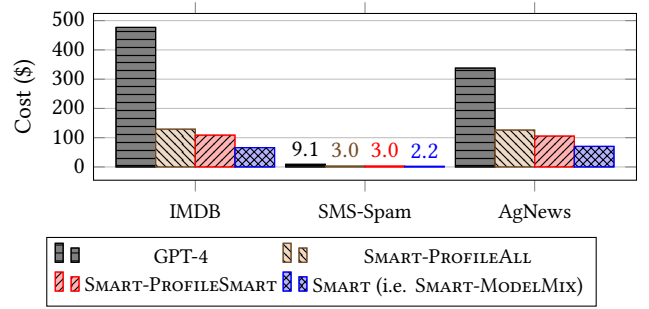

\subsection{Experimental Setup}
\label{sec:exp_setup}

\textbf{Benchmarks.} We evaluate \sys using three well-established real-world datasets, summarized in Table~\ref{tbl:benchmarks}. First, the IMDB benchmark~\cite{MaasDPHNP11} is a widely recognized dataset for binary sentiment classification. It consists of 50,000 movie reviews from the Internet Movie Database (IMDB), divided into 25,000 instances for training and the rest for testing. For our evaluation, we utilize all instances from both the training and testing sets. Second, the SMS-Spam Collection benchmark~\cite{AlmeidaHY11} is a dataset collected for the purpose of spam detection in Short Message Service (SMS). It comprises a collection of messages labeled as spam or legitimate with a total of 5,574 instances. Third, the AgNews benchmark~\cite{ZhangZL15} is a dataset designed for news categorization tasks, containing 120,000 training samples and 7,600 test instances. Each entry is a news article categorized into one of four classes: World, Sports, Business, or Science/Technology. We again use all available instances for our evaluation. Using the tokenizer for GPT-4 (gpt-4-0613), on average, the numbers of tokens per instance are 293.7, 22.9, and 51.2 for IMDB, SMS-Spam, and AgNews, respectively.

\textbf{Large Language Models.} In our experiments, we employ a range of OpenAI~\cite{OpenAI} models as our available LLMs. Specifically, we use the following set of models: gpt-4-0613, gpt-3.5-turbo-instruct, gpt-3.5-turbo-1106, davinci-002, and babbage-002. The costs per 1,000 input tokens for these models are $\$0.03$, $\$0.0015$, $\$0.001$, $\$0.002$, and $\$0.0004$, respectively. We designate GPT-4 (gpt-4-0613) as the reference model for our experiments. We employ the same prompt template across all LLMs, presenting a text and requesting to classify it into one of the given labels. 

\textbf{Evaluation Setup.} To mitigate the high cost of running all instances from the benchmarks on all LLMs, we sample 100 instances from each benchmark and evaluate them using all LLMs. We then calculate the accuracy of the outputs from each LLM, checking their equivalence compared to those of GPT-4. To simulate the actual run, we create a dataset where each instance has a probability, equal to the accuracy of each LLM, of matching the output of GPT-4. This dataset maintains the same number of instances as the original. We repeat this simulation 10 times for each benchmark and report the average values unless noted otherwise. Throughout the experiments, we use a confidence level of $\gamma = 0.95$. For the accuracy constraint $\delta$, we set values ranging from 0.02 up to 0.2, with increments of 0.02. For example, an accuracy constraint of $\delta = 0.02$ implies that \sys produces outputs consistent with those of GPT-4 in 98\% of cases with a 95\% confidence level. As our primary metric, we use cost savings, calculated as $\text{cost}_{\text{before}} / \text{cost}_{\text{after}}$. For example, if GPT-4 incurs a cost of \$300 and \sys incurs a cost of \$30 on the same benchmark, this indicates a cost reduction by 10$\times$. We evaluate all variants of \sys, including \sysa, \sysb, and \sysc. Given that \sysc represents the final variant of the framework, we subsequently refer to it as \sys.

\begin{figure}
    \centering
    \begin{tikzpicture}
        \begin{groupplot}[group style={group size=1 by 3, vertical sep=1.40cm}, width=8.8cm, height=3.8cm, ybar=0, xlabel={}, ylabel={Cost (\$)}, label style={font=\small, align=center}, enlarge x limits=0.1, typeset ticklabels with strut, xticklabels={0.02, 0.04, 0.06, 0.08, 0.10, 0.12, 0.14, 0.16, 0.18, 0.20}, xticklabel style={font=\small}, legend entries={GPT-4, \sysa, \sysb, \sys (i.e.\, \sysc)}, legend style={font=\small, at={(0.45,-0.41)}, anchor=north, legend columns=2}, xtick=data, ymajorgrids, ylabel near ticks, legend to name=groupPlotLegend, cycle list name=patternList]
        \nextgroupplot[title={IMDB}, bar width=4pt, ytick={0, 100, 200, 300, 400, 500}]
            \addplot table[skip first n=0, x expr=\coordindex, y index=1, col sep=comma] {plots/imdb.csv};
            \addplot table[skip first n=0, x expr=\coordindex, y index=2, col sep=comma] {plots/imdb.csv};
            \addplot table[skip first n=0, x expr=\coordindex, y index=3, col sep=comma] {plots/imdb.csv};
            \addplot table[skip first n=0, x expr=\coordindex, y index=4, col sep=comma] {plots/imdb.csv};
        \nextgroupplot[title={SMS-Spam}, bar width=4pt]
            \addplot table[skip first n=0, x expr=\coordindex, y index=1, col sep=comma] {plots/sms_spam.csv};
            \addplot table[skip first n=0, x expr=\coordindex, y index=2, col sep=comma] {plots/sms_spam.csv};
            \addplot table[skip first n=0, x expr=\coordindex, y index=3, col sep=comma] {plots/sms_spam.csv};
            \addplot table[skip first n=0, x expr=\coordindex, y index=4, col sep=comma] {plots/sms_spam.csv};
        \nextgroupplot[title={AgNews}, bar width=4pt, xlabel={Accuracy Constraint $\delta$}]
            \addplot table[skip first n=0, x expr=\coordindex, y index=1, col sep=comma] {plots/ag_news.csv};
            \addplot table[skip first n=0, x expr=\coordindex, y index=2, col sep=comma] {plots/ag_news.csv};
            \addplot table[skip first n=0, x expr=\coordindex, y index=3, col sep=comma] {plots/ag_news.csv};
            \addplot table[skip first n=0, x expr=\coordindex, y index=4, col sep=comma] {plots/ag_news.csv};
        \end{groupplot}
    \end{tikzpicture}
    \ref{groupPlotLegend}
    \caption{Costs of GPT-4 and all variants of \sys with varying accuracy constraint $\delta$.}
    \label{fig:detailed}
\end{figure}
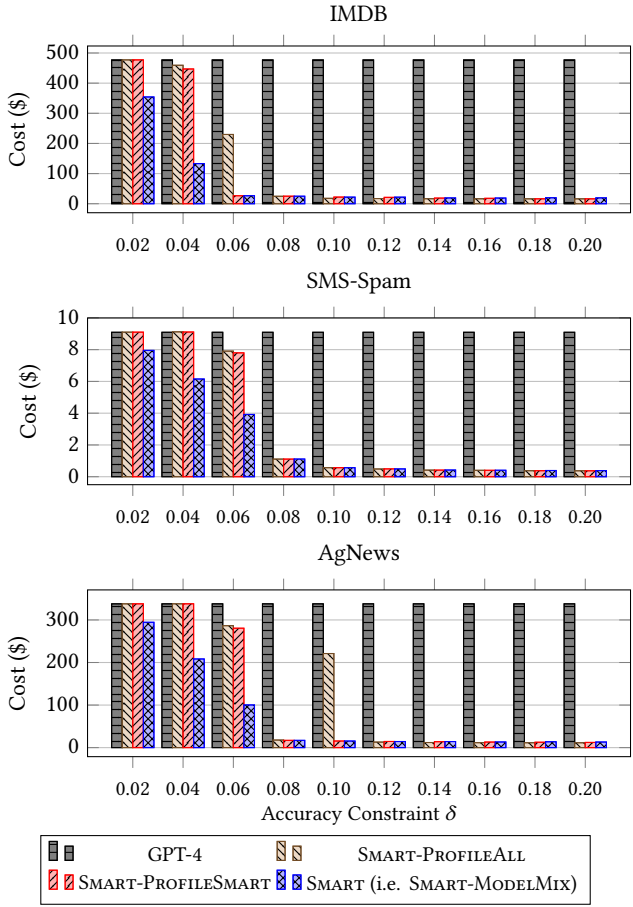

\begin{table}
    \centering
    \small
    \caption{Minimum and mean accuracy (\%) of \sys across 10 runs for varying accuracy thresholds $1 - \delta$, along with the number of runs failing to meet the constraint.\label{tbl:accuracy}}
    \begin{tabular}{l|rrr|rrr|rrr}
        \toprule[1pt]
        & \multicolumn{3}{c|}{\textbf{IMDB}} & \multicolumn{3}{c|}{\textbf{SMS-Spam}} & \multicolumn{3}{c}{\textbf{AgNews}} \\
        $\mathbf{1 - \delta}$ & Min & Avg & \#F  & Min & Avg & \#F & Min & Avg & \#F \\
        \midrule[1pt]
        0.98 & 98.4 & 98.8 & - & 98.6 & 99.1 & - & 98.8 & 99.1 & - \\
        0.96 & 96.6 & 96.9 & - & 96.8 & 97.9 & - & 96.8 & 97.6 & - \\
        0.94 & 95.7 & 96.0 & - & 94.0 & 95.6 & 1/10 & 94.2 & 95.3 & - \\
        0.92 & 95.1 & 95.8 & - & 93.2 & 94.0 & - & 93.7 & 93.9 & - \\
        0.90 & 93.9 & 94.8 & - & 92.8 & 93.5 & - & 88.7 & 93.3 & 1/10 \\
        0.88 & 93.4 & 94.8 & - & 87.5 & 92.5 & 1/10 & 89.5 & 92.3 & - \\
        0.86 & 91.3 & 93.2 & - & 85.5 & 92.5 & 1/10 & 87.4 & 91.8 & - \\
        0.84 & 89.2 & 91.0 & - & 83.5 & 90.9 & 1/10 & 85.4 & 89.7 & - \\
        0.82 & 92.6 & 93.7 & - & 86.6 & 91.4 & - & 88.6 & 91.2 & - \\
        0.80 & 90.3 & 91.5 & - & 84.5 & 89.8 & - & 86.5 & 89.1 & - \\
        \bottomrule[1pt]
    \end{tabular}
\end{table}

\subsection{Experimental Results}
\label{sec:exp_result}

Figure~\ref{fig:aggregated} displays the average costs, measured in dollars, incurred by GPT-4 and \sys variants across all benchmarks, aggregated over all accuracy constraints. All \sys variants outperform GPT-4 in terms of cost across all the benchmarks. \sys significantly reduces expenses compared to GPT-4, achieving cost savings of 7.2$\times$, 4.2$\times$, and 4.8$\times$ for the IMDB, SMS-Spam, and AgNews benchmarks, respectively. Specifically, for the IMDB benchmark, \sys demonstrates cost savings of 7.2$\times$ compared to GPT-4. Compared to the preceding variants \sysa and \sysb, \sys achieves cost savings of 2.0$\times$ and 1.6$\times$, respectively. For the SMS-Spam benchmark, \sys shows cost reductions of 4.2$\times$ compared to GPT-4, and 1.4$\times$ compared to both \sysa and \sysb. Lastly, for the AgNews benchmark, \sys achieves cost savings of 4.8$\times$, 1.8$\times$, and 1.5$\times$ compared to GPT-4, \sysa, and \sysb.

Figure~\ref{fig:detailed} illustrates the costs associated with using GPT-4 and \sys variants across all benchmarks, under varying user-defined accuracy constraints. As the accuracy constraint $\delta$ is relaxed, \sys achieves greater cost savings. With an accuracy constraint of $\delta = 0.02$, \sys realizes cost savings of 1.3$\times$, 1.1$\times$, and 1.1$\times$ compared to GPT-4 for the IMDB, SMS-Spam, and AgNews benchmarks, respectively. When the value of the accuracy constraint is increased to $\delta = 0.06$, \sys achieves savings of 18.0$\times$, 2.3$\times$, and 3.4$\times$ for IMDB, SMS-Spam, and AgNews, respectively. Further increasing the accuracy constraint to $\delta = 0.1$, \sys obtains even higher savings of 21.7$\times$, 16.0$\times$, and 21.8$\times$, respectively. This trend demonstrates the efficiency of \sys in reducing more costs as users impose less demanding accuracy requirements.

\begin{figure}
    \centering
    \begin{tikzpicture}
        \begin{groupplot}[group style={group size=1 by 3, vertical sep=1.40cm}, ybar stacked, width=8cm, height=3.8cm, scaled ticks=false, xlabel={}, ylabel={\#Processed}, label style={font=\small, align=center}, enlarge x limits=0.1, typeset ticklabels with strut, xticklabels={0.02, 0.04, 0.06, 0.08, 0.10, 0.12, 0.14, 0.16, 0.18, 0.20}, xticklabel style={font=\small, align=center}, legend entries={gpt-4-0613, gpt-3.5-turbo-instruct, gpt-3.5-turbo-1106, davinci-002, babbage-002}, legend style={font=\small, at={(0.45,-0.41)}, anchor=north, legend columns=3}, xtick=data, ymajorgrids, ylabel near ticks, legend to name=groupPlotLegend, cycle list name=patternList2]
        \nextgroupplot[title={IMDB}, bar width=12pt]
            \addplot table[x expr=\coordindex, y index=1, col sep=comma] {plots/breakdown.csv};
            \addplot table[x expr=\coordindex, y index=2, col sep=comma] {plots/breakdown.csv};
            \addplot table[x expr=\coordindex, y index=3, col sep=comma] {plots/breakdown.csv};
            \addplot table[x expr=\coordindex, y index=4, col sep=comma] {plots/breakdown.csv};
            \addplot table[x expr=\coordindex, y index=5, col sep=comma] {plots/breakdown.csv};
        \nextgroupplot[title={SMS-Spam}, bar width=12pt]
            \addplot table[x expr=\coordindex, y index=6, col sep=comma] {plots/breakdown.csv};
            \addplot table[x expr=\coordindex, y index=7, col sep=comma] {plots/breakdown.csv};
            \addplot table[x expr=\coordindex, y index=8, col sep=comma] {plots/breakdown.csv};
            \addplot table[x expr=\coordindex, y index=9, col sep=comma] {plots/breakdown.csv};
            \addplot table[x expr=\coordindex, y index=10, col sep=comma] {plots/breakdown.csv};
        \nextgroupplot[title={AgNews}, bar width=12pt, xlabel={Accuracy Constraint $\delta$}, y tick label style={/pgf/number format/fixed, /pgf/number format/fixed zerofill, /pgf/number format/precision=0}]
            \addplot table[x expr=\coordindex, y index=11, col sep=comma] {plots/breakdown.csv};
            \addplot table[x expr=\coordindex, y index=12, col sep=comma] {plots/breakdown.csv};
            \addplot table[x expr=\coordindex, y index=13, col sep=comma] {plots/breakdown.csv};
            \addplot table[x expr=\coordindex, y index=14, col sep=comma] {plots/breakdown.csv};
            \addplot table[x expr=\coordindex, y index=15, col sep=comma] {plots/breakdown.csv};
        \end{groupplot}
    \end{tikzpicture}
    \ref{groupPlotLegend}
    \caption{Breakdown of \sys in terms of the number of instances processed by each LLM. The total number of instances processed by LLMs may exceed the number of instances in the dataset due to profiling, where the same item is processed using multiple LLMs.}
    \label{fig:breakdown_process}
\end{figure}
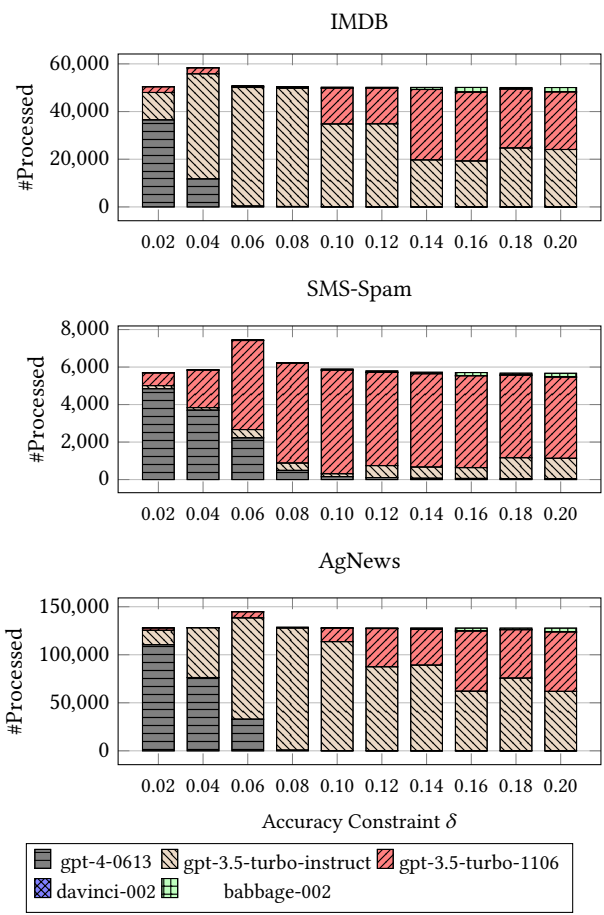

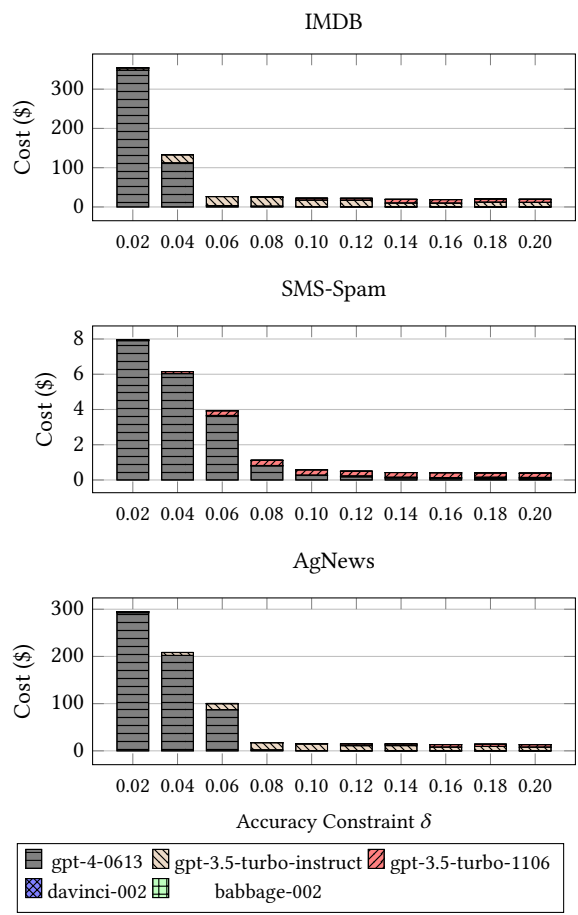
\begin{figure}
    \centering
    \begin{tikzpicture}
        \begin{groupplot}[group style={group size=1 by 3, vertical sep=1.40cm}, ybar stacked, width=8cm, height=3.8cm, scaled ticks=false, xlabel={}, ylabel={Cost (\$)}, label style={font=\small, align=center}, enlarge x limits=0.1, typeset ticklabels with strut, xticklabels={0.02, 0.04, 0.06, 0.08, 0.10, 0.12, 0.14, 0.16, 0.18, 0.20}, xticklabel style={font=\small, align=center}, legend entries={gpt-4-0613, gpt-3.5-turbo-instruct, gpt-3.5-turbo-1106, davinci-002, babbage-002}, legend style={font=\small, at={(0.45,-0.41)}, anchor=north, legend columns=3}, xtick=data, ymajorgrids, ylabel near ticks, legend to name=groupPlotLegend, cycle list name=patternList2]
        \nextgroupplot[title={IMDB}, bar width=12pt]
            \addplot table[x expr=\coordindex, y index=16, col sep=comma] {plots/breakdown.csv};
            \addplot table[x expr=\coordindex, y index=17, col sep=comma] {plots/breakdown.csv};
            \addplot table[x expr=\coordindex, y index=18, col sep=comma] {plots/breakdown.csv};
            \addplot table[x expr=\coordindex, y index=19, col sep=comma] {plots/breakdown.csv};
            \addplot table[x expr=\coordindex, y index=20, col sep=comma] {plots/breakdown.csv};
        \nextgroupplot[title={SMS-Spam}, bar width=12pt]
            \addplot table[x expr=\coordindex, y index=21, col sep=comma] {plots/breakdown.csv};
            \addplot table[x expr=\coordindex, y index=22, col sep=comma] {plots/breakdown.csv};
            \addplot table[x expr=\coordindex, y index=23, col sep=comma] {plots/breakdown.csv};
            \addplot table[x expr=\coordindex, y index=24, col sep=comma] {plots/breakdown.csv};
            \addplot table[x expr=\coordindex, y index=25, col sep=comma] {plots/breakdown.csv};
        \nextgroupplot[title={AgNews}, bar width=12pt, xlabel={Accuracy Constraint $\delta$}, y tick label style={/pgf/number format/fixed, /pgf/number format/fixed zerofill, /pgf/number format/precision=0}]
            \addplot table[x expr=\coordindex, y index=26, col sep=comma] {plots/breakdown.csv};
            \addplot table[x expr=\coordindex, y index=27, col sep=comma] {plots/breakdown.csv};
            \addplot table[x expr=\coordindex, y index=28, col sep=comma] {plots/breakdown.csv};
            \addplot table[x expr=\coordindex, y index=29, col sep=comma] {plots/breakdown.csv};
            \addplot table[x expr=\coordindex, y index=30, col sep=comma] {plots/breakdown.csv};
        \end{groupplot}
    \end{tikzpicture}
    \ref{groupPlotLegend}
    \caption{Breakdown of \sys in terms of the cost per LLM.}
    \label{fig:breakdown_cost}
\end{figure}

In Figure~\ref{fig:detailed}, the costs associated with \sysa under accuracy constraints of $\delta=0.08$ and $\delta=0.1$ on the AgNews benchmark presents an intriguing pattern. Contrary to expectations, as the constraint is relaxed from $\delta=0.08$ to $\delta=0.1$, the cost of \sysa does not decrease but instead increases. This anomaly can be attributed to the true accuracy of gpt-3.5-turbo-1106 being very close to the accuracy threshold $1 - \delta = 0.9$. When the true accuracy of an LLM lies near the accuracy threshold, the profiling phase of \sysa struggles to conclusively accept or reject the LLM as satisfying the accuracy requirement. Consequently, \sysa incurs additional costs from continuous profiling without meeting the termination criterion. In contrast, \sysb (and \sys) avoid such extra profiling costs by employing a smart termination criterion that evaluates the expected cost savings from additional profiling. Consequently, \sysa achieves cost savings of 1.5$\times$ compared to GPT-4, whereas \sysb realizes cost savings of 21.7$\times$.

Table~\ref{tbl:accuracy} presents the minimum and mean accuracy (\%) of \sys across the ten runs for each benchmark and accuracy constraint. If the minimum accuracy meets or exceeds the accuracy threshold, \sys produces outputs that satisfy the equivalence guarantees with respect to the reference model, GPT-4, in all ten runs. Additionally, Table~\ref{tbl:accuracy} details the number of runs, out of ten, in which the outputs of \sys fail to meet the accuracy constraint. In most cases, the outputs of \sys comply with the accuracy constraint for all ten runs. Even in the five scenarios where the minimum accuracy falls short of the threshold, only one run out of ten fails to meet the constraint. Across all benchmarks and accuracy constraints, merely 5 out of the 300 runs violate the accuracy constraint, corresponding to 1.6\% of cases. This rate is consistent with the confidence level $\gamma = 0.95$ used in our experiments.

Figure~\ref{fig:breakdown_process} provides a detailed view of \sys in terms of the number of instances processed by each LLM. As the accuracy constraint $\delta$ is relaxed, \sys seamlessly transitions to utilizing LLMs that, while less accurate, are more cost-efficient. For instance, when the accuracy constraint is stringent at $\delta = 0.02$, GPT-4 is responsible for the majority of processing across all benchmarks. This aligns with expectations as the outputs of \sys may differ from those of GPT-4 in no more than 2\% of instances. On the other hand, with a more relaxed accuracy constraint of $\delta = 0.1$, the processing portion of GPT-4 drops significantly to 0.2\%, 2.6\%, and 0.2\% for the IMDB, SMS-Spam, AgNews benchmarks, respectively. A key advantage of \sys is its ability to dynamically determine how many instances each LLM should process, while providing accuracy guarantees as per the user-specified accuracy constraint $\delta$ and confidence level $\gamma$. Figure~\ref{fig:breakdown_cost} similarly presents a breakdown of \sys, as shown in Figure~\ref{fig:breakdown_process}, but focuses on the cost per LLM. Although the number of processed items retains a similar value, the overall cost decreases as the accuracy constraint $\delta$ is relaxed, owing to the utilization of more cost-efficient LLMs.

\begin{figure}
    \centering
    \begin{tikzpicture}
        \begin{groupplot}[group style={group size=1 by 1, vertical sep=1.40cm}, width=8.4cm, height=3.8cm, ybar=0, xlabel={}, ylabel={Cost (\$)}, label style={font=\small, align=center}, enlarge x limits=0.1, typeset ticklabels with strut, xticklabels={ID0 \\ 0.88 \\ 0.88 \\ 0.88, ID1 \\ 0.90 \\ 0.88 \\ 0.88, ID2 \\ 0.90 \\ 0.90 \\ 0.88, ID3 \\ 0.90 \\ 0.90 \\ 0.90, ID4 \\ 0.92 \\ 0.88 \\ 0.88, ID5 \\ 0.92 \\ 0.90 \\ 0.88, ID6 \\ 0.92 \\ 0.90 \\ 0.90, ID7 \\ 0.92 \\ 0.92 \\ 0.88, ID8 \\ 0.92 \\ 0.92 \\ 0.90, ID9 \\ 0.92 \\ 0.92 \\ 0.92}, xticklabel style={font=\small, align=center}, legend entries={GPT-4, \sysa, \sysb, \sys (i.e.\, \sysc)}, legend style={font=\small, at={(0.45,-0.41)}, anchor=north, legend columns=2}, xtick=data, ymajorgrids, ylabel near ticks, legend to name=groupPlotLegend, cycle list name=patternList]
        \nextgroupplot[title={}, bar width=4pt, xlabel={\\ \\ \\ \\ Accuracy Levels of LLMs}]
            \addplot table[skip first n=0, x expr=\coordindex, y index=1, col sep=comma] {plots/simulation.csv};
            \addplot table[skip first n=0, x expr=\coordindex, y index=2, col sep=comma] {plots/simulation.csv};
            \addplot table[skip first n=0, x expr=\coordindex, y index=3, col sep=comma] {plots/simulation.csv};
            \addplot table[skip first n=0, x expr=\coordindex, y index=4, col sep=comma] {plots/simulation.csv};
        \end{groupplot}
        \node[] at (-0.55,-0.79) {\footnotesize gpt-3.5-instruct:};
        \node[] at (-0.4,-1.15) {\footnotesize gpt-3.5-1106:};
        \node[] at (-0.4,-1.51) {\footnotesize babbage-002:};
    \end{tikzpicture}
    \ref{groupPlotLegend}
    \caption{Costs of GPT-4 and all variants of \sys for an accuracy constraint of $\delta=0.1$ across varying accuracy levels of LLMs.}
    \label{fig:simulation}
\end{figure}
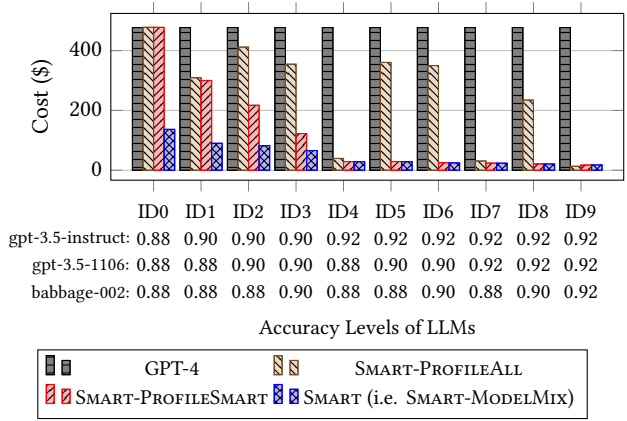

\begin{figure}
    \centering
    \begin{tikzpicture}
        \begin{groupplot}[group style={group size=3 by 1, vertical sep=1.40cm, horizontal sep=0.1cm}, width=3.95cm, height=4.5cm, scaled ticks=false, ymin=-0.1, extra y ticks={0.9}, extra y tick labels={\textbf{\textcolor{red}{0.9}}}, xlabel={\#Items}, ylabel={Confidence Interval}, label style={font=\small}, xticklabel style={font=\small}, legend entries={gpt-3.5-turbo-instruct(lower), gpt-3.5-turbo-instruct(upper), gpt-3.5-turbo-1106(lower), gpt-3.5-turbo-1106(upper), davinci-002(lower), davinci-002(upper), babbage-002(lower), babbage-002(upper)}, legend style={font=\small, at={(0.45,-0.41)}, anchor=north, legend columns=2}, xmajorgrids, ymajorgrids, ylabel near ticks, legend to name=groupPlotLegend]
        \nextgroupplot[title={IMDB}, xmin=-10, xmax=120]
            \addplot[mark=none, red, forget plot] coordinates {(-10, 0.9) (120, 0.9)};
            \addplot[mark=x, black] table[x index=0, y index=7, col sep=comma] {plots/ci_imdb.csv};
            \addplot[mark=x, black] table[x index=0, y index=8, col sep=comma] {plots/ci_imdb.csv};
            \addplot[mark=o, brown!80!black, mark size=1.5pt] table[x index=0, y index=5, col sep=comma] {plots/ci_imdb.csv};
            \addplot[mark=o, brown!80!black, mark size=1.5pt] table[x index=0, y index=6, col sep=comma] {plots/ci_imdb.csv};
            \addplot[mark=diamond, green!50!black] table[x index=0, y index=3, col sep=comma] {plots/ci_imdb.csv};
            \addplot[mark=diamond, green!50!black] table[x index=0, y index=4, col sep=comma] {plots/ci_imdb.csv};
            \addplot[mark=square, blue] table[x index=0, y index=1, col sep=comma] {plots/ci_imdb.csv};
            \addplot[mark=square, blue] table[x index=0, y index=2, col sep=comma] {plots/ci_imdb.csv};
        \nextgroupplot[title={SMS-Spam}, ylabel={}, yticklabels={}, extra y tick labels={}, xmin=-5, xmax=65]
            \addplot[mark=none, red, forget plot] coordinates {(-5, 0.9) (65, 0.9)};
            \addplot[mark=x, black] table[x index=0, y index=7, col sep=comma] {plots/ci_sms_spam.csv};
            \addplot[mark=x, black] table[x index=0, y index=8, col sep=comma] {plots/ci_sms_spam.csv};
            \addplot[mark=o, brown!80!black, mark size=1.5pt] table[x index=0, y index=5, col sep=comma] {plots/ci_sms_spam.csv};
            \addplot[mark=o, brown!80!black, mark size=1.5pt] table[x index=0, y index=6, col sep=comma] {plots/ci_sms_spam.csv};
            \addplot[mark=diamond, green!50!black] table[x index=0, y index=3, col sep=comma] {plots/ci_sms_spam.csv};
            \addplot[mark=diamond, green!50!black] table[x index=0, y index=4, col sep=comma] {plots/ci_sms_spam.csv};
            \addplot[mark=square, blue] table[x index=0, y index=1, col sep=comma] {plots/ci_sms_spam.csv};
            \addplot[mark=square, blue] table[x index=0, y index=2, col sep=comma] {plots/ci_sms_spam.csv};
        \nextgroupplot[title={AgNews}, ylabel={}, yticklabels={}, extra y tick labels={}, xmin=-30, xmax=370]
            \addplot[mark=none, red, forget plot] coordinates {(-30, 0.9) (370, 0.9)};
            \addplot[mark=x, black] table[x index=0, y index=7, col sep=comma] {plots/ci_ag_news.csv};
            \addplot[mark=x, black] table[x index=0, y index=8, col sep=comma] {plots/ci_ag_news.csv};
            \addplot[mark=o, brown!80!black, mark size=1.5pt] table[x index=0, y index=5, col sep=comma] {plots/ci_ag_news.csv};
            \addplot[mark=o, brown!80!black, mark size=1.5pt] table[x index=0, y index=6, col sep=comma] {plots/ci_ag_news.csv};
            \addplot[mark=diamond, green!50!black] table[x index=0, y index=3, col sep=comma] {plots/ci_ag_news.csv};
            \addplot[mark=diamond, green!50!black] table[x index=0, y index=4, col sep=comma] {plots/ci_ag_news.csv};
            \addplot[mark=square, blue] table[x index=0, y index=1, col sep=comma] {plots/ci_ag_news.csv};
            \addplot[mark=square, blue] table[x index=0, y index=2, col sep=comma] {plots/ci_ag_news.csv};
        \end{groupplot}
    \end{tikzpicture}
    \ref{groupPlotLegend}
    \caption{Lower and upper bounds of confidence intervals on LLM accuracy during the profiling phase.}
    \label{fig:confidence_interval}
\end{figure}
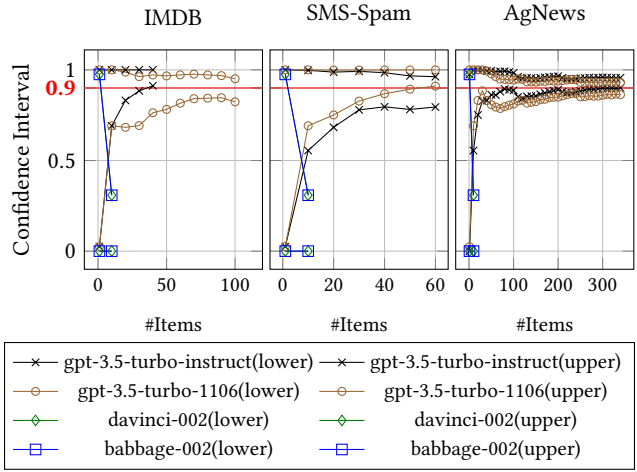

In Figure~\ref{fig:simulation}, we evaluate \sys using a synthetic benchmark designed to vary the accuracy levels of available LLMs. We first create a dataset that replicates the data properties of the IMDB benchmark, such as the number of instances and tokens. We then set the accuracy constraint to $\delta = 0.1$ and assign the accuracy of the underlying LLMs (except for the reference LLM, GPT-4) to one of the following values: $\{0.88, 0.90, 0.92\}$. These values simulate scenarios in which the accuracy of an LLM is below, at, or above the user-defined accuracy threshold of $1 - \delta = 0.9$. This experiment aims to demonstrate the adaptability of \sys across various combinations of LLM accuracy levels. The LLMs used in the experiment include gpt-4-0613, gpt-3.5-turbo-instruct, gpt-3.5-turbo-1106, and babbage-002. To simplify the analysis, we reduce the number of LLMs involved by excluding davinci-002. As shown in Figure~\ref{fig:simulation}, \sys outperforms both GPT-4 as well as other variants, \sysa and \sysb, across various scenarios. Specifically, \sys is the only method to achieve cost reduction when all LLMs have an accuracy level below the accuracy threshold of $1 - \delta = 0.9$ (ID0). In scenarios where the accuracy of an LLM is exactly the same as the specified accuracy threshold (ID2, ID3, ID5, ID6, ID8), making it difficult to either accept or reject the LLM as satisfying the accuracy constraint, \sysa suffers from additional profiling overheads. In contrast, \sysb (and by extension, \sys) can terminate profiling early if further profiling is expected to be wasteful based on cost estimations. \sys shows superior performance over \sysb, particularly in scenarios when the accuracy of all LLMs is lower than or equal to the accuracy threshold (ID0, ID1, ID2, ID3). \sys is able to leverage these cheaper LLMs, even when they fall short of meeting the accuracy guarantees directly, in conjunction with the more accurate but more expensive GPT-4.

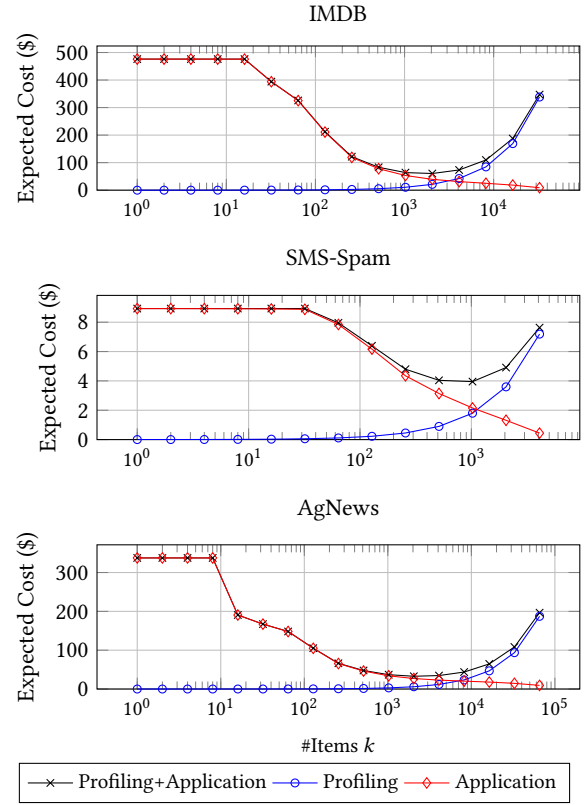
\begin{figure}
    \centering
    \begin{tikzpicture}
        \begin{groupplot}[group style={group size=1 by 3, vertical sep=1.40cm}, width=8cm, height=3.5cm, scaled ticks=false, ymin=0,
        xmode=log, xtick={1,10,100,1000,10000,100000},
        xlabel={}, ylabel={Expected Cost (\$)}, label style={font=\small}, xticklabel style={font=\small}, legend entries={Profiling+Application, Profiling, Application}, legend style={font=\small, at={(0.45,-0.41)}, anchor=north, legend columns=3}, xmajorgrids, ymajorgrids, ylabel near ticks, legend to name=groupPlotLegend]
        \nextgroupplot[title={IMDB}, ytick={0, 100, 200, 300, 400, 500}]
            \addplot[mark=x, black] table[x index=0, y index=1, col sep=comma] {plots/cost_imdb.csv};
            \addplot[mark=o, blue, mark size=1.5pt] table[x index=0, y index=2, col sep=comma] {plots/cost_imdb.csv};
            \addplot[mark=diamond, red] table[x index=0, y index=3, col sep=comma] {plots/cost_imdb.csv};
        \nextgroupplot[title={SMS-Spam}]
            \addplot[mark=x, black] table[x index=0, y index=1, col sep=comma] {plots/cost_sms_spam.csv};
            \addplot[mark=o, blue, mark size=1.5pt] table[x index=0, y index=2, col sep=comma] {plots/cost_sms_spam.csv};
            \addplot[mark=diamond, red] table[x index=0, y index=3, col sep=comma] {plots/cost_sms_spam.csv};
        \nextgroupplot[title={AgNews}, xlabel={\#Items $k$}]
            \addplot[mark=x, black] table[x index=0, y index=1, col sep=comma] {plots/cost_ag_news.csv};
            \addplot[mark=o, blue, mark size=1.5pt] table[x index=0, y index=2, col sep=comma] {plots/cost_ag_news.csv};
            \addplot[mark=diamond, red] table[x index=0, y index=3, col sep=comma] {plots/cost_ag_news.csv};
        \end{groupplot}
    \end{tikzpicture}
    \ref{groupPlotLegend}
    \caption{Expected costs (including costs of the application phase) associated with profiling exactly $k$ additional items with exponentially increasing $k$.}
    \label{fig:expected_cost}
\end{figure}

\red{
Figure~\ref{fig:confidence_interval} illustrates the lower and upper bounds of confidence intervals on the accuracy of each LLM during the profiling phase. We present the case where the accuracy threshold is set to $1 - \delta = 0.9$, which is marked in red on the y-axis. As more items are profiled and additional information is collected, the confidence intervals narrow down further. If the upper confidence interval bound of an LLM falls below the accuracy threshold (as illustrated for babbage-002 on IMDB), the LLM is deemed \texttt{Invalid}, and this LLM is no longer profiled. Conversely, if the lower bound of the interval surpasses the accuracy threshold, the LLM is considered \texttt{Valid} (as demonstrated for gpt-3.5-turbo-instruct on IMDB). Profiling concludes upon identifying the most cost-efficient LLM with \texttt{Valid} status (for SMS-Spam) or when further profiling is expected to be wasteful (for IMDB and AgNews). The number of items processed for profiling is significantly less compared to the total number of input instances across all benchmarks.
}

\begin{figure}
    \centering
    \begin{tikzpicture}
        \begin{groupplot}[group style={group size=2 by 1, vertical sep=1.40cm, horizontal sep=0.7cm}, width=4.8cm, height=3.5cm, scaled ticks=false, ymin=0,
        xmode=log, xtick={1,10,100,1000,10000,100000},
        xlabel={\#Items $k$}, ylabel={Expected Cost (\$)}, label style={font=\small}, xticklabel style={font=\small}, legend entries={Profiling+Application, Profiling, Application}, legend style={font=\small, at={(0.45,-0.41)}, anchor=north, legend columns=3}, xmajorgrids, ymajorgrids, ylabel near ticks, legend to name=groupPlotLegend]
        \nextgroupplot[title={IMDB}]
            \addplot[mark=x, black] table[x index=0, y index=1, col sep=comma] {plots/cost_inverse_imdb.csv};
            \addplot[mark=o, blue, mark size=1.5pt] table[x index=0, y index=2, col sep=comma] {plots/cost_inverse_imdb.csv};
            \addplot[mark=diamond, red] table[x index=0, y index=3, col sep=comma] {plots/cost_inverse_imdb.csv};
        \nextgroupplot[title={AgNews}, ylabel={}]
            \addplot[mark=x, black] table[x index=0, y index=1, col sep=comma] {plots/cost_inverse_ag_news.csv};
            \addplot[mark=o, blue, mark size=1.5pt] table[x index=0, y index=2, col sep=comma] {plots/cost_inverse_ag_news.csv};
            \addplot[mark=diamond, red] table[x index=0, y index=3, col sep=comma] {plots/cost_inverse_ag_news.csv};
        \end{groupplot}
    \end{tikzpicture}
    \ref{groupPlotLegend}
    \caption{Expected costs associated with profiling exactly $k$ additional items with exponentially increasing $k$. Scenarios where further profiling is expected to be wasteful based on cost estimations.}
    \label{fig:expected_cost_inverse}
\end{figure}
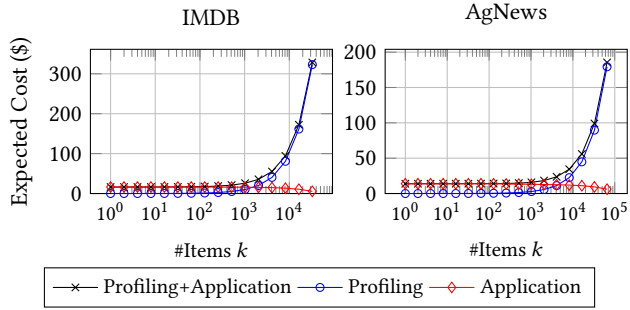

Figure~\ref{fig:expected_cost} illustrates the expected costs associated with profiling exactly $k$ additional items. \red{It is important to note that these expected costs include the impact of profiling on the anticipated costs of the subsequent application phase. Thus, we present the expected costs for both the profiling and application phases, as well as the total costs.} \sys enables the early termination of profiling, leveraging the cost estimations introduced in Section~\ref{sec:sysb}. The figure demonstrates that, as $k$ increases exponentially, the expected costs have a pattern as anticipated. \red{As the number of profiled items increases, the expected cost for profiling also increases. Conversely, the expected cost of the application phase decreases as we gather more accurate information about the accuracy of LLMs. Thus, the overall expected cost exhibits the following pattern.} Initially, increasing the number of profiled items reduces the associated expected costs until reaching a specific point. Beyond this point, further increases in the number of profiled items lead to increased expected costs. This pattern suggests the existence of an optimal $k$, at which the balance between the costs of profiling and the savings gained during the application phase is optimal. The benefits of profiling surpass its overheads up to a certain point, at which sufficient information about LLMs has been acquired. Beyond this point, profiling additional items yields little new information, making the costs of further profiling outweigh its benefits. As an additional observation, the figure shows that the exponential scheme for increasing the value of $k$ adequately encompasses the possible values of expected costs associated with varying $k$. This experiment features the estimated expected costs of profiling more items after \sys has profiled 100 items. \red{Here, we apply an accuracy constraint of $\delta = 0.1$. Additionally, Figure~\ref{fig:expected_cost_inverse} again illustrates the expected costs of profiling $k$ more items, highlighting cases where further profiling is anticipated to be wasteful. In this figure, the costs of profiling largely outweigh the cost savings during the application phase for all values of $k$. That is, the expected cost has the lowest value when there is no additional profiling. In such scenarios, \sys opt to terminate profiling early and proceed to the application phase. A plot for SMS-Spam is not presented, as \sys decides to continue profiling in all cases under the current setting.}

\section{Related Work}
\label{sec:related}

The Transformer architecture~\cite{VaswaniSPUJGKP17} revolutionized the field of natural language processing (NLP) with significant improvements across multiple tasks. Building on the Transformer architecture, a series of large language models (LLMs) emerged, demonstrating state-of-the-art performance on various NLP tasks in both zero-shot~\cite{KojimaGRMI22} and few-shot settings~\cite{BrownMRSKDNSSAA20, Wei0SBIXCLZ22}. The number of parameters in LLMs has significantly increased along with their inference costs~\cite{abs-2305-05176}, as models with more parameters have shown better performance~\cite{abs-2001-08361, abs-2203-15556}. As a result, there has been growing interest in enhancing the efficiency of LLM inference~\cite{abs-2312-15234}, with strategies including cascade models~\cite{abs-2310-03094, abs-2305-05176, abs-2310-03046, JitkrittumGMNRK23}, early exit mechanisms~\cite{SchwartzSSDS20, SchusterFG0B0TM22}, caching results~\cite{abs-2306-02003}, batch prompting~\cite{abs-2309-00384, abs-2301-08721}, compressed LLMs~\cite{abs-2305-11186}, and distributed inference~\cite{abs-2305-05920}.


Model cascading is closely relevant to \sys. Specifically, FrugalGPT~\cite{abs-2305-05176} is a learning-based framework that utilizes LLM cascading to process NLP tasks within a cost budget. This cascade relies on a sequence of LLMs with varying costs. For any given query, if the first LLM returns a response that is deemed reliable (according to a scoring function), FrugalGPT outputs its response. If not, FrugalGPT escalates the query to the next LLM in the sequence. To learn the score threshold for each LLM, it requires ground truth labels for the task at hand. On the other hand, \sys considers the scenario where there are no ground truth labels and employs a systematic approach to decide how many items to process using the reference LLM. FrugalGPT, as well as other model cascading frameworks, reduces cost by dynamically adjusting compute per input complexity. In contrast, \sys is designed for scenarios where a substantial number of input instances are evaluated for a common question. Hence, \sys models the difficulty of the question and selects appropriate LLMs at the task level. Above all, \sys provides equivalence guarantees on the generated outputs.

\sys is related to approximate query processing~\cite{ChaudhuriDK17} in that it enables users to trade result precision for reduced cost. A significant body of prior work~\cite{Olken1995, AgarwalPMIMS12, AgarwalMPMMS13, KandulaSVOGCD16, NirkhiwaleDJ13, ParkMSW18, JoT20} allows users to specify an accuracy constraint in addition to their query. The IFocus algorithm~\cite{KimBPIMR15} is relevant as it regularly updates confidence intervals to generate approximate visualizations of large datasets. Unlike these approaches, which are primarily focused on relational data, \sys is specifically designed to process a large number of NLP task instances.
\section{Conclusion}
\label{sec:conclusion}

\sys is a framework designed to minimize the cost of large-scale LLM inference for NLP tasks, while providing guarantees on the quality of its outputs. Through a novel profiling scheme and the strategic use of multiple LLMs, \sys enables users to achieve significant cost savings safely, while ensuring high-quality results. Using OpenAI LLMs, our experimental evaluation across three real-world datasets demonstrates its ability to dramatically reduce costs compared to GPT-4.


\bibliographystyle{ACM-Reference-Format}
\bibliography{library}

\end{document}